\documentclass[11pt]{article}

\usepackage[final]{acl}

\usepackage{times}
\usepackage{latexsym}

\usepackage[T1]{fontenc}

\usepackage[utf8]{inputenc}

\usepackage{microtype}
\usepackage{hyperref}
\usepackage{url}
\usepackage{booktabs}
\usepackage{xspace}
\usepackage{graphicx}
\usepackage{textcomp}
\usepackage{multicol}
\usepackage{multirow}
\usepackage{enumitem}
\usepackage{pifont}
\usepackage{amsmath}
\usepackage{subfigure}
\usepackage{float}
\usepackage{tikz}
\usepackage{subcaption}
\usepackage{lineno}
\usepackage{makecell}
\usepackage{CJKutf8}
\usepackage{dblfloatfix}

\usepackage{inconsolata}

\usepackage{graphicx}

\definecolor{darkblue}{rgb}{0, 0, 0.5}
\hypersetup{colorlinks=true, citecolor=darkblue, linkcolor=darkblue, urlcolor=darkblue}

\def \name{\textsc{EMCEE}\xspace}
\definecolor{mycolor}{RGB}{33, 95, 154}
\definecolor{custom_red}{RGB}{228, 54, 54}

\title{\name{}: Improving Multilingual Capability of LLMs via Bridging Knowledge and Reasoning with Extracted Synthetic Multilingual Context}

\author{
  Hamin Koo \quad
  Jaehyung Kim \\
  Yonsei University \\
  \texttt{\{hamin2065,jaehyungk\}@yonsei.ac.kr}
}

\begin{document}
\maketitle

\begin{abstract}
Large Language Models (LLMs) have achieved impressive progress across a wide range of tasks, yet their heavy reliance on English-centric training data leads to significant performance degradation in non-English languages.
While existing multilingual prompting methods emphasize reformulating queries into English or enhancing reasoning capabilities, they often fail to incorporate the language- and culture-specific grounding that is essential for some queries.
To address this limitation, we propose \name{} (\textbf{E}xtracting synthetic \textbf{M}ultilingual \textbf{C}ont\textbf{e}xt and m\textbf{e}rging), a simple yet effective framework that enhances the multilingual capabilities of LLMs by explicitly extracting and utilizing query-relevant knowledge from the LLM itself.
In particular, \name{} first extracts synthetic context to uncover latent, language-specific knowledge encoded within the LLM, and then dynamically merges this contextual insight with reasoning-oriented outputs through a judgment-based selection mechanism.
Extensive experiments on four multilingual benchmarks covering diverse languages and tasks demonstrate that \name{} consistently outperforms prior approaches, achieving an average relative improvement of 16.4\% overall and 31.7\% in low-resource languages.\footnote{Code: \url{https://github.com/hamin2065/EMCEE}}
\end{abstract}
\section{Introduction}

\begin{figure}[ht]
\centering
\includegraphics[width=\columnwidth]{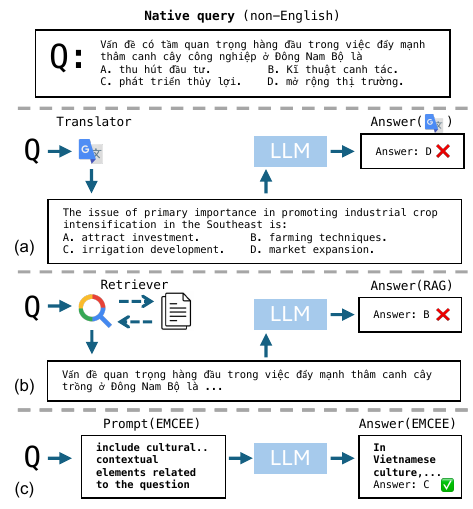}
\vspace{-0.3in}
\caption{
\textbf{ 
Different multilingual prompting methods.} 
Given a Vietnamese query from social science category in M3-Exam \citep{zhang2023m3exam}, \textbf{(a)} Translating the query into English using the external translator results in an incorrect answer. \textbf{(b)} Even with retrieval-augmented generation (Google Custom Search API), the model remains incorrect. \textbf{(c)} However, with \name{} (Ours) prompt that extracts relevant context from LLM itself, the LLM successfully produces the correct answer.
}\label{fig:comparison}
\vspace{-0.25in}
\end{figure}

Large Language Models (LLMs) have demonstrated remarkable capabilities across a wide range of applications, from natural language understanding to generation \citep{brown2020language,touvron2023llama,team2024codegemma}.
Among the numerous factors contributing to LLM performance, the quality and quantity of pre-training data stand out as particularly critical \citep{kaplan2020scaling, hoffmann2022training, gunasekar2023textbooks}.
However, given that most pre-training data are in English \citep{ouyang2022training, touvron2023llama}, LLMs inevitably exhibit an English-centric bias.
As a result, their performance often deteriorates when handling non-English queries \citep{shi2022language}.
As the use of LLMs in the real-world rapidly increases, this disparity highlights the urgent need for developing LLMs or methods that can perform effectively across diverse languages, especially those with limited resources.

\begin{figure*}[ht]
\vspace{-0.15in}
\centering
\includegraphics[width=1.0\textwidth]{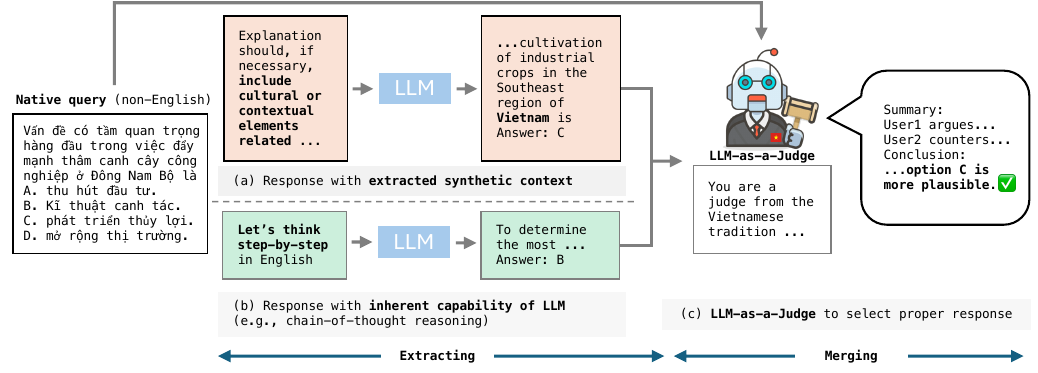}
\vspace{-0.1in}
\caption{
\textbf{Overview of \name{}.} (a) LLM receives a non-English (native) query along with an instruction to \textit{extract relevant synthetic context}, producing a context-enriched response. (b) In parallel, LLM generates a \textit{reasoning-focused response} using only its inherent knowledge, without additional context. (c) An \textit{LLM-as-a-Judge} module then compares the two responses and selects the final answer based on contextual relevance and reasoning adequacy.
}
\vspace{-0.15in}
\label{fig:method}
\end{figure*}

Addressing multilingual queries with LLMs is particularly challenging because they often involve heterogeneous requirements that demand different strategies. 
For example, queries in domains such as mathematics and natural science primarily depend on reasoning skills, which are less sensitive to the query language itself. 
Existing multilingual prompting approaches have mostly targeted this case by translating non-English queries into English \citep{liu2024translation} or by inducing intermediate reasoning in English \citep{huang2023not}. 
On the other hand, queries in domains such as language or social science rely heavily on knowledge grounded in specific countries or cultures; in such cases, incorporating appropriate knowledge is more crucial than applying abstract reasoning. 
However, this remains a significant challenge, as LLMs often struggle to acquire and utilize language- and culture-specific information due to the limited representation of such non-English data in pre-training corpora; 
yet, how to effectively handle such knowledge-intensive multilingual queries remains under-explored.
Furthermore, since both reasoning-oriented and knowledge-intensive queries are often mixed in the real-world, an effective system should be able to answer both cases simultaneously, introducing additional complexity.\looseness=-1

\textbf{Contribution.} To fill this gap, we propose a new prompting-based framework called \name{} (\textbf{E}xtracting synthetic \textbf{M}ultilingual \textbf{C}ont\textbf{e}xt and m\textbf{e}rging), which enables LLMs to better handle heterogeneous multilingual queries. 
Specifically, \name{} consists of two key components: (1) an \textit{explicit extraction step} that generates synthetic context to surface query-relevant knowledge from the LLM, and (2) a \textit{merging step} that adaptively integrates this knowledge with reasoning. 
The extraction step is motivated by \citet{sun2023recitation}, which demonstrates that explicitly eliciting hidden knowledge into textual form is often more effective than relying on the model to recall it implicitly. 
Building on this insight, \name{} enriches the context for answering multilingual queries by incorporating synthetic data, which is especially crucial for knowledge-intensive cases. 
Unlike retrieval-augmented generation, this approach neither depends on external retrievers nor databases; 
instead, our framework fully leverages the LLM's own internal knowledge representation to generate synthetic, language-specific context.
Simultaneously, \name{} derives an alternative reasoning-focused response, which tends to be more effective for reasoning-oriented queries. 
Finally, \name{} employs an LLM-as-a-Judge \citep{zheng2023judging} to merge the LLM’s reasoning across the two candidate responses, enabling the system to determine whether a knowledge-grounded or reasoning-oriented path is more suitable for each query.

\vspace{0.1em}
We show the effectiveness of \name{} through extensive evaluations on four multilingual datasets. 
These datasets cover various languages and tasks such as multiple-choice question answering, text generation, language understanding, and reasoning. 
For example, compared to a method that uses native query with native instructions, \name{} achieves an average relative improvement of 16.4\% across all datasets, with a notable improvement of 31.7\% in low-resource languages.
These results suggest that \name{} effectively captures and leverages cultural knowledge, addressing a key limitation of prior work.
In addition, we find that \name{} is compatible with various state-of-the-art LLMs, including API-based and Open-source ones. 
Overall, these findings underscore the importance of cultural knowledge extraction for non-English queries and the need for continued research in multilingual prompting that incorporates cultural awareness.
\section{Enhancing Multilingual Capability of LLMs with Extracted Synthetic Context}

\subsection{Preliminary}

We consider the problem of generating an accurate response for a non-English input query, referred to as a \textit{native query}.
As most LLMs are trained primarily on English data, their performance on non-English queries is often limited \citep{shi2022language}.
To mitigate this, \textit{multilingual prompting} aims to reformulate the input prompt so that it better aligns with the model's inherent knowledge.
For instance, \citet{liu2024translation} translates the native query into English, and \citet{huang2023not} further proposes to construct a mixed input prompt composed of native queries and English instructions that induce chain-of-thought (CoT) reasoning.
However, these methods are limited in capturing language- and culture-specific contexts, including social conventions and background knowledge embedded in the native language, that is critical for knowledge-intensive queries.\looseness=-1

\subsection{\name{}: Extract Synthetic Multilingual Context and Merge with Reasoning}
To address these limitations, our key idea is to extract query-relevant information from the LLM itself and use it as an additional context to improve the response quality. 
As illustrated in Figure \ref{fig:method}, \name{} follows a two-stage framework designed to enhance the model's ability to answer native queries by incorporating relevant contextual knowledge. 
The first stage, \textit{extraction}, elicits from the LLM the implicit background knowledge required to answer the query.
The second stage, \textit{merging}, integrates this knowledge with the model's reasoning to produce a more accurate final response.

\begin{figure}[ht]
\vspace{-0.1in}
\begin{center}
    {
    \subfigure[Prompt and LLM response for extracting]
        {
        \includegraphics[width=\columnwidth]{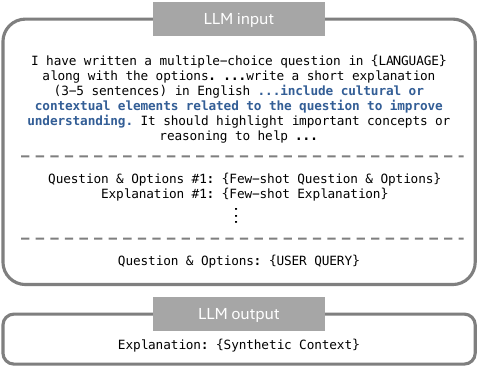}
        \label{fig:prompt_extract}
        }
    \subfigure[Prompt and LLM response for merging]
        {
        \includegraphics[width=\columnwidth]{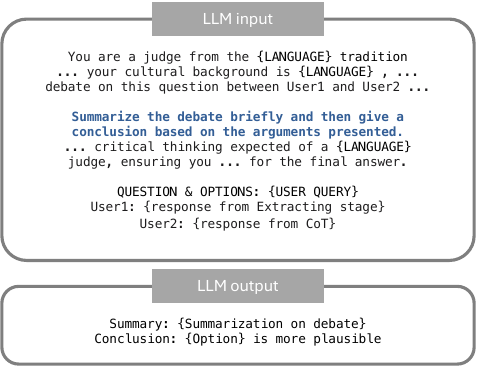}
        \label{fig:prompt_emulsify}
        } 
    }
\vspace{-0.1in}
\caption{\textbf{Input prompt and LLM response example.} These figures illustrates the overview of input prompt and response of LLM during (a) Extracting and (b) Merging processes, respectively.
}
\label{fig:prompt}
\end{center}
\vspace{-0.3in}
\end{figure}

\paragraph{Extracting relevant context from LLM.}
The goal of this stage is to extract query-relevant knowledge from the LLM in the form of \textit{synthetic context}, supplementing information that the model often overlooks in direct query answering.
To this end, we prompt the LLM to generate background knowledge necessary for answering the native query.
In line with prior work, studies have shown that explicitly eliciting knowledge encoded in model parameters can improve downstream performance across diverse tasks \citep{sun2023recitation, wang2023query2doc}.
Following this perspective, our approach exploits the internal knowledge embedded within the LLM's parameters, without relying on external retrieval or databases.
Concretely, we design an instruction template that guides the model to extract relevant information, typically within 3–5 sentences (see Figure \ref{fig:prompt} (a)). The extracted context is then concatenated with the native query to create an augmented input, enabling the LLM to generate responses that are better informed and contextually grounded.
To further improve the quality of the extracted context, we employ a few-shot prompting setup with exemplars demonstrating how to extract essential background knowledge.
This overall setup enables the model to surface knowledge spanning cultural, historical, and domain-specific dimensions.
Additionally, to leverage LLMs' well-established ability to follow English instructions, we conduct this extraction step with English instructions (see Appendix~\ref{sec:ext_native} for more details).

\paragraph{Merging synthetic context with reasoning.}
While extracting synthetic context enables the LLM to better handle knowledge-intensive multilingual queries, the degree of reliance on such knowledge varies widely across tasks.
For instance, in the M3-Exam dataset \citep{zhang2023m3exam}, questions from social science or language subjects benefit from contextual grounding, whereas those from mathematics or natural-science primarily rely on logical reasoning.
To accommodate this diversity within a unified framework, we generate an additional reasoning-focused response and use it with the context-enriched one to obtain the final output.

\vspace{1.5mm}
Specifically, we adopt CoT prompting with English instruction to obtain an additional response focusing on the LLM's inherent reasoning capabilities. 
We then employ an LLM-as-a-Judge framework \citep{zheng2023judging} to select the most appropriate response, as illustrated in Figure \ref{fig:method}.
The judge model is instructed to consider the linguistic and contextual background of the query, such as local customs, social norms, or discipline-specific reasoning patterns, thereby selecting the response that best fits the query's context (see Figure \ref{fig:prompt} (b)).
By selectively combining contextual and reasoning-oriented responses, \name{} merges the LLM’s reasoning to evaluate and choose the response best suited to each query, ensuring robust performance across diverse multilingual settings.
While it is also possible to prompt the LLM to choose a strategy from the query alone, we observe that our judgment-based merging---where the LLM compares and reasons over the generated responses---yields better performance, likely because it avoids the additional decision-stage errors (see Table~\ref{tab:main}).
\vspace{2mm}
\section{Experiments}\label{sec:4}

\subsection{Setups}\label{sec:4.1}

\paragraph{Datasets.}
We evaluate the effectiveness of \name{} across four tasks: multiple-choice question answering (MCQA), text generation, language understanding, and reasoning. 
For each task, we utilize the following benchmarks to ensure a comprehensive, multilingual evaluation: 
(1) \textit{M3-Exam} \citep{zhang2023m3exam} is an exam-style dataset collected from diverse countries. 
We exclude image-based questions and use only text data to assess culturally and linguistically varied test items. 
In order to conduct a fair evaluation, we remove every question that contains background knowledge and keep only those that can be answered directly from the prompt.
(2) \textit{MKQA} \citep{longpre-etal-2021-mkqa} is a generation-based evaluation benchmark for open-domain question answering across 26 languages. Following \citet{huang2023not}, we select 9 languages.
(3) \textit{XNLI} \citep{conneau-etal-2018-xnli} evaluates logical entailment understanding across 14 languages. 
(4) \textit{XCOPA} \citep{ponti-etal-2020-xcopa} evaluates commonsense reasoning in 11 low-resource languages. 
Across all datasets, English is excluded to prioritize multilingual generalization. 
To ensure a balanced cross-lingual analysis, we randomly sample 100 examples per language for MKQA, XNLI, and XCOPA, while using the entire filtered M3-Exam. 
Overall, the combined datasets cover a span of 24 languages. More details about the experimental datasets are provided in Appendix~\ref{supp:dataset}.

\paragraph{Metrics.} 
We use accuracy for M3-Exam, XNLI, and XCOPA since these tasks involve selecting a single correct answer from predefined options. 
For MKQA, we use a span-based F1 score, which measures the token-level overlap between the predicted answer and the ground truth. 
In addition, to assess model performance comprehensively, we report our results in three different ways: (1) averaging across all languages (\textit{All}), (2) using only high-resource languages (\textit{High}), and (3) using only low-resource languages (\textit{Low}). 
Since no exact criteria exist for defining high- and low-resource languages, we use a dataset-specific empirical approach; specifically, we compute the average performance on native-language queries for each dataset and use this as a threshold, \textit{e.g.}, where languages scoring above the threshold are considered high-resource.

\begin{table*}[t]
\vspace{-0.1in}
\caption{\textbf{Main result.} Test performance of GPT-4o-mini over different multilingual prompting methods on four different tasks. Results are measured using accuracy for M3-Exam, XNLI, and XCOPA. F1-score is used for MKQA. The best and second best scores are highlighted in \textbf{bold} and \underline{underline}, respectively. \textit{All}, \textit{High}, and \textit{Low} categories are determined based on performance of \textsc{Native-Basic}: the top half of languages are labeled high-resource, and the bottom half low-resource. The numbers in parentheses indicate the number of evaluated languages.}
\vspace{-0.1in}
\begin{center}
\resizebox{0.9\textwidth}{!}{%
\begin{tabular}{@{}l|ccc|ccc|ccc|ccc@{}}
\toprule
{Datasets ($\rightarrow$)} & \multicolumn{3}{c}{{M3-Exam}} & \multicolumn{3}{c}{{MKQA}} & \multicolumn{3}{c}{{XNLI}} & \multicolumn{3}{c}{{XCOPA}} \\
{Methods ($\downarrow$)} & All & High & Low & All & High & Low & All & High & Low & All & High & Low  \\ 
& (8) & (4) & (4) & (9) & (5) & (4) & (14) & (7) & (7) & (11) & (6) & (5)  \\ 

\midrule
  \textsc{Native-Basic} & 
  65.2 & 72.7  & 57.7 & 
  44.1 & 48.6 & 38.5 & 
  66.2 & 74.0 & 58.4 & 
  79.3 & 94.2 & 61.4 \\ 
\textsc{Eng-Basic} & 
      64.5 & 72.7 & 56.4 &  
      48.1 & 49.8 & 46.0 & 
      71.4 & \underline{74.6} & 68.1 & 
      55.3 & 57.3 & 52.8      \\
\textsc{Trans-Google} & 
    59.9 & 70.1 & 49.8 & 
    50.3 & 50.4 & 50.2 & 
    70.5 & 73.4 & 67.6 & 
    66.0 & 65.2 & 67.3 \\
\textsc{XLT}& 
    70.4 & 76.9 & 63.8 & 
    \underline{51.1} & 50.9 & \underline{51.5} & 
    72.6 & 74.3  & 71.0 & 
    \underline{91.1} & 95.8 & \underline{85.4}      \\
\textsc{Native-Cot}   & 
    70.4 & 76.5 & 64.4 & 
    45.5 & 49.2 & 40.9 & 
    68.4 & 70.1  & 66.6 & 
    90.0 & \underline{96.3} & 82.4     \\
\textsc{Eng-Cot}
    & 74.6 & 81.8 & 67.3 & 
    49.4 & 49.5 & 49.3 & 
    \underline{73.2} & 73.7 & \underline{72.7} & 
    90.5 & 96.0 & 83.8     \\ 
\textsc{RAG (Native)}
    & 68.7 & 80.4 & 57.0 &
    42.5 & 45.3  & 39.0 &
    67.4 & 69.9 & 64.9 &
    83.6 & 92.3 & 73.2 \\
\textsc{RAG (Eng)}
    & 72.1 & \textbf{84.0} & 63.9 &
    44.7 & 44.9 & 44.5 &
    70.4 & 71.1 & 69.7 &
    87.9 & 94.0 & 80.6 \\
\midrule 
\name{} \textsc{(Route)}
    & \underline{76.2} & 83.1 & \underline{69.2} &
    50.8 & \underline{51.6}  & 49.8 &
    73.1 & 73.9 & 72.3 &
    90.5 & 96.0 & 83.8 \\
  \name{} (Ours) & 
  \textbf{77.4} &\underline{83.3} & \textbf{71.5} & 
  \textbf{52.3} & \textbf{52.2} & \textbf{52.4} & 
  \textbf{74.3} & \textbf{74.7} & \textbf{73.9} & 
  \textbf{92.0} & \textbf{96.8} & \textbf{86.2} \\ 
\bottomrule
\end{tabular}%
}
\end{center}
\vspace{-0.15in}
\label{tab:main}
\end{table*}

\paragraph{Baselines.}
Following previous works \citep{huang2023not, liu2024translation}, we consider baselines that account for variations in instruction language, chain-of-thought (CoT) reasoning, and translation.
We adopt a zero-shot approach for all methods, given that our chosen models are already fine-tuned to follow instructions.
(1) \textsc{Native-Basic}: All instructions and questions are written in the native language without reasoning steps. This setup measures performance entirely within the native language context. 
(2) \textsc{Eng-Basic}: This setting is derived from \textsc{Native-Basic}, except that the instructions are provided in English while the questions remain in the native language. This highlights how partial English usage can affect performance. 
(3) {\textsc{XLT}} \citep{huang2023not}: XLT is a prompting method that improves the multilingual capability of LLMs. It instructs the model to translate the question into English and subsequently solve the problem step-by-step in English. 
(4) {\textsc{Trans-Google}} \citep{liu2024translation}: All instructions and questions are translated (via Google Translate API) into English before being processed by the model. This baseline shows how external machine translation impacts overall performance. 
(5) \textsc{Native-Cot}: Both instructions and CoT reasoning are provided in the native language, allowing the model to “think” in that language.
(6) {\textsc{Eng-Cot}}: Both instructions and CoT reasoning are carried out in English, assessing how full English prompts with detailed reasoning steps influence results.
(7) \textsc{RAG (Native)}: This setting extends \textsc{Native-Basic} by appending retrieved passages (in the native language) obtained via the Google Custom Search API as an external context.
(8) \textsc{RAG (Eng)}: This variant combines the \textsc{Eng-Basic} setup with retrieval-augmented generation, where English instructions are used alongside retrieved passages.
Specific prompt templates of these settings are presented in Table~\ref{tab:prompt_template}.

\paragraph{Implementation details.}
We employ GPT-4o-mini (\texttt{gpt-4o-mini-2024-07-18}) \citep{openai2024gpt4omini}.
To mitigate the influence of random sampling, we strictly set the temperature to 0.0 for API-based models and use greedy decoding for Llama models.
See more details in Appendix \ref{supp:details}.

\begin{figure*}[t]
\centering
\vspace{-0.2in}
\includegraphics[width=0.85\textwidth]{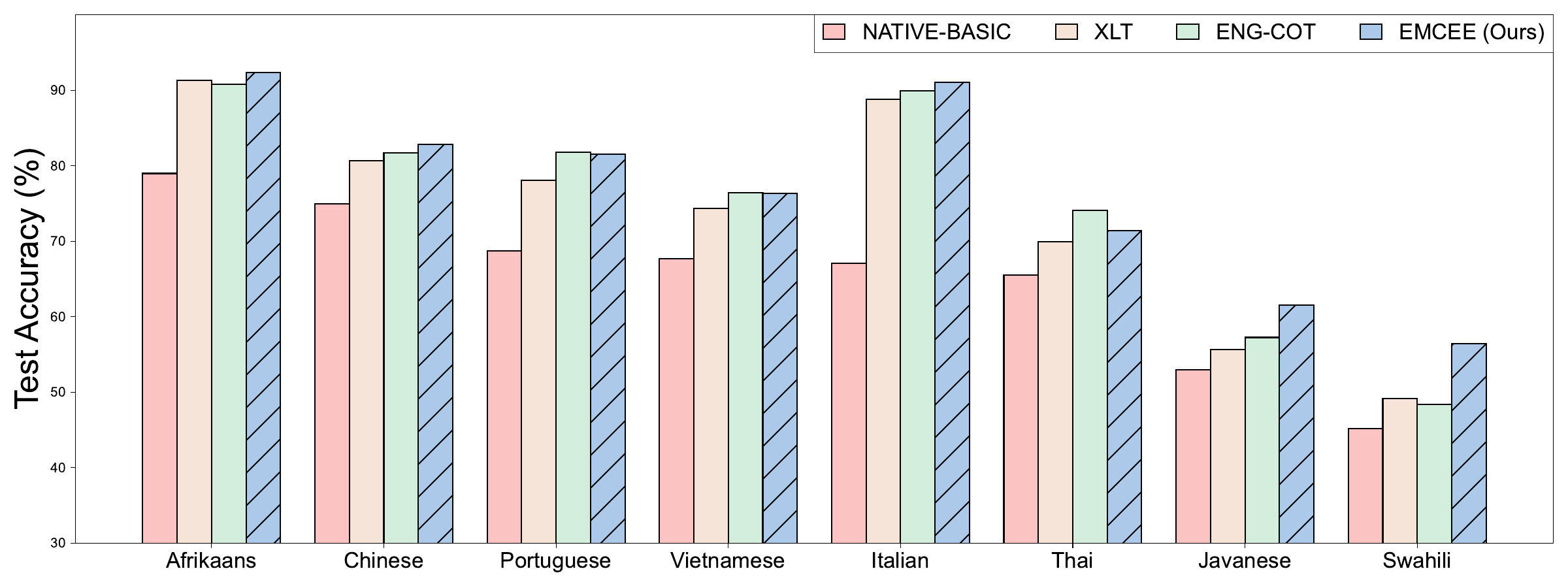}
\vspace{-0.1in}
\caption{\textbf{Overall language-wise improvement.} Test accuracy of GPT-4o-mini over four different multilingual prompting methods on M3-Exam. More results on other datasets and LLMs are presented in Appendix \ref{supp:analyses}.
}
\vspace{-0.1in}
\label{fig:language_acc}
\end{figure*}

\subsection{Main results}
Table~\ref{tab:main} presents the results of our experiments on four multilingual benchmarks using GPT-4o-mini.
Overall, \name{} consistently surpasses all non-RAG baselines across datasets, achieving a 16.4\% average relative improvement over the basic prompting method, \textsc{Native-Basic}, and further improving upon non-RAG baselines by 3.2\% (\textsc{Eng-CoT}) and 3.9\% (\textsc{XLT}). 
Notably, \name{} shows the largest gains in low-resource languages—a result especially important for multilingual modeling.
Compared to \textsc{Native-Basic}, it improves by 23.7\% on M3-Exam, 36.1\% on MKQA, 27.7\% on XNLI, and 40.4\% on XCOPA, averaging a 32.0\% gain. 
\name{} also exceeds \textsc{Eng-CoT} and \textsc{XLT} by 4.0\% and 4.4\%, respectively. 
In addition, Figure~\ref{fig:language_acc} further confirms that our framework delivers substantial boosts for low-resource languages.

We also compare against retrieval-augmented generation (RAG) methods (\textsc{RAG-Native} and \textsc{RAG-Eng}) that utilize external context, which is different from how \name{} uses synthetic context.
Although these RAG methods offer minor gains over simpler baselines, their performance remains lower than that of reasoning-oriented approaches such as \textsc{Eng-CoT} and \textsc{XLT}, and far behind \name{} in most cases. 
For example, on M3-Exam, \textsc{RAG-Eng} reaches 72.1\% compared to 74.6\% for \textsc{Eng-CoT} and 77.4\% for \name{}, with similar patterns across MKQA, XNLI, and XCOPA. 
This suggests that external retrieval offers limited benefit, while \name{}’s extraction of \textit{synthetic multilingual context} from the model’s own knowledge yields richer, query-aligned context.
This highlights that \name{} fully exploits the internal knowledge and reasoning capabilities of the underlying LLM itself, without relying on any external retrieval modules or auxiliary corpora, thereby making it a purely self-contained prompting framework.\looseness=-1

Lastly, we further include a routing variant, \name{}~(Route), which lets the LLM select between extraction and reasoning pathways based on language characteristics of the query alone instead of our merging step. 
While it achieves strong results overall, it occasionally underperforms the unified \name{} due to routing errors or suboptimal pathway selection. 
This implies that merging the LLM’s reasoning over both extraction and reasoning strategies is more robust than deciding on a single strategy based solely on the query. \looseness=-1

\subsection{More analyses}
In this section, we conduct additional analyses of \name{} on the M3-Exam dataset, with GPT-4o-mini as the default LLM unless otherwise specified.

\begin{table}[t]
\caption{\textbf{Ablation study of \name{}.} Test accuracy of GPT-4o-mini on M3-Exam with different configurations of the proposed components in \name{}. CoT, ExT, MeR are abbreviations of Chain-of-Thought, ExTracting, and MeRging, respectively.}
\begin{center}
\resizebox{1.0\linewidth}{!}{
\begin{tabular}{l|ccc|c}
\toprule
Methods & CoT & ExT & MeR & All / High / Low  \\ \midrule

\textsc{Native-Basic}                    & \textcolor{red}{\ding{55}}           &  \textcolor{red}{\ding{55}}                & \textcolor{red}{\ding{55}} & 65.2 / 72.7 / 57.7 \\
\textsc{Eng-CoT}                 & \textcolor{green}{\ding{51}}           &  \textcolor{red}{\ding{55}}               & \textcolor{red}{\ding{55}} & 74.6 / 81.8 / 67.3 \\
                 & \textcolor{red}{\ding{55}}           &  \textcolor{green}{\ding{51}}               & \textcolor{red}{\ding{55}} & 74.7 / 82.0 / 67.5 \\
                  & \textcolor{green}{\ding{51}}           &  \textcolor{red}{\ding{55}}                  & \textcolor{green}{\ding{51}}  & 75.2 / 83.4 / 67.1 \\ \midrule
\name{}                 & \textcolor{green}{\ding{51}}           &  \textcolor{green}{\ding{51}}                  & \textcolor{green}{\ding{51}}    & 77.4 / 83.3 / 71.5 \\ 
\bottomrule
\end{tabular}
}
\end{center}
\vspace{-0.2in}
\label{tab:ablation}
\end{table}
\paragraph{Ablation study.} 
We conduct an ablation study to evaluate the contributions of the three core components of \name{}: (1) Chain-of-Thought (CoT) reasoning, (2) synthetic knowledge extraction (ExT), and (3) our merging (MeR) mechanism guided by an LLM-as-a-Judge framework. 
The results are summarized in Table~\ref{tab:ablation}.
First, by comparing the results on the first and third rows, one can verify that using the extracted knowledge as an additional input context leads to improved performance. 
Next, in the fourth row, we consider a hybrid setup for the ablation. 
We generate an additional CoT response using a positive temperature and apply the same \name{} merging step between this sampled response and the original CoT-generated response.
In other words, the response based on extracted information is substituted with another CoT-generated response. 
As shown in the improvement on low-resource languages (67.1\% $\rightarrow$ 71.5\%) between the fourth and fifth rows, the impact of merging two responses from different knowledge sources is evident. \looseness=-1
\begin{figure*}[t]
    \centering
    \vspace{-0.15in}
    \small
    \begin{tabular}{p{0.45\linewidth} p{0.45\linewidth}}
    \toprule 
    \textbf{(1) Native query}\vspace{0.01in} \newline  Pagupon Iku Jenenge Kandhang ... \newline
    A. Pitik\quad B. Macan\quad C. Jaran\quad D. Dara&
    \textbf{(2) Translated query}\vspace{0.01in}\newline Pagupon is the name of the cage for ... \newline
    A. Chicken\quad B. Tiger\quad C. Horse\quad D. Dove \\
    \midrule 
    \multicolumn{2}{p{0.95\linewidth}}{\textbf{(3) Incorrect responses by \textsc{Eng-CoT}:} 1. A. Pitik ... "chicken" in Javanese ... referred to as a "kandhang pitik." ... 4. D. Dara ... means "dove" in Javanese ... referred to as a "kandhang dara." ... \textcolor{custom_red}{\textbf{the most appropriate answer would be related to domesticated animals. The correct answer is A. Pitik, as it directly refers to a barn for chickens.}}} \\
    \midrule 
    \multicolumn{2}{p{0.95\linewidth}}{\textbf{(4) Correct response by {Extracting}:} ... In Javanese, the term "kandhang" refers to a cage or pen for animals. Among the options provided, "Dara" refers to a pigeon or dove in Javanese. \textcolor{mycolor}{\textbf{Therefore, the correct answer is D, Dara, as it directly relates to the type of animal that would be housed in a "kandhang" specifically designed for pigeons.}} The other options refer to different animals that would have their own respective enclosures.} \\
    \midrule 
    \multicolumn{2}{p{0.95\linewidth}}{\textbf{(5) \name{}'s response by Merging (3) \& (4) :} User1 argues that the question is about the name of a barn for domesticated animals, specifically identifying "kandhang pitik" for chickens as the most appropriate answer. User2 counters that "pagupon" specifically refers to pigeons, and therefore "kandhang dara" is the correct term for a pigeon shelter. \textcolor{mycolor}{\textbf{Conclusion: Given that "pagupon" directly translates to "pigeon" in Javanese, and the question specifically asks for the name of the shelter for that animal, [option] D is more plausible.}}} \\
    \bottomrule
    \end{tabular}
    \caption{\textbf{Qualitative example from M3-Exam.} This figure shows: (1) native Javanese questions and options, (2) the translated one by {GPT-4o}, (3) incorrect responses produced by \textsc{Eng-CoT}, (4) correct responses obtained through our cultural knowledge extraction, and (5) an \name{} response that integrates and compares the outputs from (3) and (4). More qualitative examples are provided in Appendix \ref{supp:examples}.}
    \label{fig:examples}
    \vspace{-0.1in}
\end{figure*}

\begin{table}[t]
    \caption{\textbf{Compatibility across various LLMs.} Average accuracy of three different LLMs on M3-Exam for all languages (All). The full results are presented in Table \ref{tab:diff_llm_full}. The best and second best scores are highlighted in \textbf{bold} and \underline{underline}, respectively.}
    \vspace{0.0in}
	\begin{center}
\resizebox{1.0\linewidth}{!}{
	\begin{tabular}{l|ccc}
 		\toprule
		Methods & GPT-4o & Claude-Haiku & Llama-3.1-8B \\ \midrule
		\textsc{Native-Basic} & 78.1 & 67.4 & 49.8 \\   
		\textsc{Trans-Google} & 68.4 & 60.1 & \underline{53.9} \\   
		\textsc{XLT} & 81.8 & 74.0 & 25.8 \\   
		\textsc{Native-CoT} & \underline{83.9} & 74.8 & 36.8 \\   
		\textsc{Eng-CoT} & 82.4 & \underline{75.4} & 40.4 \\   	
        \midrule
            \name{} (Ours) & \textbf{85.7} & \textbf{75.6} & \textbf{56.9} \\ \bottomrule
	\end{tabular}
    }
    \end{center}
    \label{table:diff_llms}
    \vspace{-0.2in}
\end{table}
\paragraph{Compatibility across LLMs.}  We perform additional experiments to showcase the efficacy of \name{} across various LLMs. 
In particular, we evaluate GPT-4o (\texttt{gpt-4o-2024-08-06}), Claude-3.5-Haiku (\texttt{claude-3-5-haiku-20241022}) \citep{anthropic2024claude35haiku}, and Llama-3.1-8B-Instruct \citep{dubey2024llama} on M3-Exam with results summarized in Table \ref{table:diff_llms}.
Compared to other baselines, \name{} consistently achieves the highest performance. 
Notably, while results on Llama-3.1-8B-Instruct show that \textsc{XLT}, \textsc{Native-CoT}, and \textsc{Eng-CoT} fail to outperform \textsc{Native-Basic}, our method maintains the highest scores. 
Overall, in GPT-4o, our method achieves 8.9\% relative improvement over \textsc{Native-Basic} on average, while the improvements were 10.8\% for Claude and 12.5\% for Llama.
These findings indicate the effectiveness of our approach across diverse architectures and resource conditions.
Full results are provided in Table~\ref{tab:diff_llm_full}.
\begin{table}[t]
\vspace{-0.05in}
\small
\caption{\textbf{Test accuracy on GlobalOpinionQA.} Best and second best scores are highlighted in \textbf{bold} and \underline{underline}, respectively.}
\vspace{-0.1in}
\begin{center}
\resizebox{0.8\linewidth}{!}{%
\begin{tabular}{l|ccc}
\toprule
Methods & All & High & Low \\ \midrule
  \textsc{Native-Basic} & 65.3 & 76.8 & 53.7  \\
      \textsc{Eng-Basic}     & \underline{68.6} & \underline{78.9} & \underline{58.3} \\
      \textsc{Trans-Google} & 68.5 & \textbf{79.7} & 57.4 \\
      \textsc{XLT}          & 55.2 & 66.0 & 44.5  \\
      \textsc{Native-Cot}   & 63.2 & 73.9 & 52.4  \\
      \textsc{Eng-Cot}      & {65.4} & {74.9} & {53.9}  \\ \midrule 
  \name{} (Ours) & 
  \textbf{69.0} &
  {77.6} & \textbf{60.4} \\ \bottomrule
\end{tabular}%
}
\vspace{-0.25in}
\end{center}
\label{tab:table_global}
\end{table}

\vspace{-0.05in}
\paragraph{Cultural alignment on subjective QA.}
Here, we further investigate whether \name{} is capable of steering LLM responses to align with specific groups of users, which has become increasingly important as LLMs are applied to subjective tasks.
For this evaluation, we use GlobalOpinionQA \citep{durmus2024towards}—a dataset comprising questions and answers from cross-national surveys designed to capture diverse global opinions.
Since the dataset was originally created in English, we employ GPT-4o to translate it into our target languages and filter it to include only countries with more than 40 samples.
Our filtered dataset covers data from over 30 countries (see Appendix~\ref{supp:dataset} for full details), totaling 2,609 samples.
We configure the LLM-as-a-Judge by \textit{country} rather than by \textit{language}, since the dataset contains overlapping languages (\textit{e.g.}, Spanish for both Spain and Peru), allowing us to distinguish between culturally distinct regions.
As shown in Table~\ref{tab:table_global}, \name{} achieves the highest overall accuracy and particularly strong performance in low-resource countries, highlighting its robustness in culturally diverse contexts.
Although \textsc{Trans-Google} slightly outperforms \name{} in high-resource settings---probably because the data were originally in English---these results confirm the effectiveness of \name{} in steering LLM outputs toward region-specific perspectives by leveraging extracted synthetic knowledge.

\vspace{-0.05in}
\begin{table}[t]
\vspace{-0.05in}
\centering
\small
\caption{\textbf{Results on Aya-8B.} Accuracy on four M3-Exam languages that are supported by multilingual model Aya-8B. Best scores are highlighted in \textbf{bold}.}
\begin{tabular}{lccccc}
\toprule
Methods & zh & it & pt & vi & Avg \\
\midrule
\textsc{Native-Basic} & 39 & 45 & 52 & 48 & 46.0 \\
\textsc{Eng-Basic} & 31 & \textbf{46} & 52 & 26 & 38.8 \\
\textsc{\name{}} & \textbf{56} & 40 & \textbf{54} & \textbf{49} & \textbf{49.8} \\
\bottomrule
\vspace{-0.25in}
\end{tabular}
\label{tab:aya_results}
\end{table}

\paragraph{Evaluation on Aya-8B.} \label{sec:aya}
To further verify the generality of our findings, we conduct additional experiments using \textsc{Aya-8B} \citep{ustun2024aya}, a model specifically optimized for multilingual understanding. 
We focus on four languages that are covered by both \textsc{Aya-8B} and the M3-Exam dataset---Chinese (zh), Italian (it), Portuguese (pt), and Vietnamese (vi)---and sample 100 queries for each language. 
As shown in Table~\ref{tab:aya_results}, \name{} improves the average performance of Aya-8B from 46.0 and 38.8 to 49.8. Gains are observed in Chinese, Portuguese, and Vietnamese, while Italian remains challenging, suggesting that \name{} remains beneficial on average even for multilingual-specialized models.

\begin{table}[t]
\centering
\small
\footnotesize 
\setlength{\tabcolsep}{3.5pt} 
\caption{\textbf{Results on stronger LLMs.} Accuracy on a subset of M3-Exam (100 samples per language) for GPT-5 and Claude-4-Sonnet. Best and second best scores are highlighted in \textbf{bold} and \underline{underline}, respectively.}
\label{tab:stronger_models}
\begin{tabular}{l ccc c ccc}
\toprule
& \multicolumn{3}{c}{\textbf{GPT-5}} & & \multicolumn{3}{c}{\textbf{Claude-4-Sonnet}} \\ 
\cmidrule(lr){2-4} \cmidrule(lr){6-8}
Methods & All & High & Low & & All & High  & Low \\ \midrule
\textsc{Native-Basic}    & \underline{74.3}         & \underline{83.8}          & \textbf{64.8} & & 79.3         & \textbf{90.0} & 68.5          \\
\textsc{Eng-CoT}         & 68.4         & 81.3          & 55.5          & & \underline{81.5}         & \underline{89.5}          & \underline{73.5}          \\
\textsc{XLT}             & 58.4         & 70.0          & 46.8          & & \textbf{82.3} & \underline{89.5}          & \textbf{75.0} \\ \midrule
\name{} \textsc{(Ours)} & \textbf{76.0} & \textbf{87.5} & \underline{64.5} & & \textbf{82.3} & \underline{89.5} & \textbf{75.0} \\ 
\bottomrule
\vspace{-0.2in}
\end{tabular}
\end{table}
\paragraph{Generalization to stronger LLMs.}
We further evaluate \name{} on GPT-5 and Claude-4-Sonnet using a subset of the M3-Exam dataset (100 per language).
As shown in Table ~\ref{tab:stronger_models}, \name{} achieves the best overall accuracy on GPT-5 (76.0), outperforming all baselines and showing strong gains on high-resource languages (87.5). 
On Claude-4-Sonnet, \name{} attains 82.3, matching the strongest baseline (XLT) while maintaining robust performance across both high- and low-resource settings.
Overall, these results indicate that \name{} generalizes well to stronger LLMs and consistently provides stable improvements over both multilingual and English-based reasoning baselines.

\begin{table}[t]
\centering
\small
\caption{\textbf{Cost analysis.} Average test accuracy, input and output token usages, and incurred cost (\$) of GPT-4o-mini on M3-Exam with other cost-intensive baseline and \name{}. The better scores are highlighted in \textbf{bold}.}
\resizebox{1.0\linewidth}{!}{
\begin{tabular}{lccc}
\toprule
Methods & Acc $\uparrow$ & Input / Output Tokens $\downarrow$ & Cost $\downarrow$ \\
\midrule
\makecell[l]{3 $\times$ \textsc{ENG-COT} \\ \& \textsc{MERGE}} & 76.9 & \textbf{282k} / 176k & \$0.149 \\
\midrule
\makecell[l]{\name{} \textsc{(Ours)}} & \textbf{78.8} & 539k / \textbf{99k} & \textbf{\$0.140} \\
\bottomrule
\end{tabular}
}
\vspace{-0.2in}
\label{tab:cost_analysis}
\end{table}
\paragraph{Cost analysis.} \label{sec:cost}
As the proposed \name{} is based on multi-step inference of LLMs, one might be concerned about its computational cost with higher token usage. 
To demonstrate that the gain from \name{} is not merely a result of increased computation, we consider other cost-intensive approaches.
Specifically, we consider a three-sample variant of the \textsc{Eng-Cot} baseline, which enhances reasoning diversity by generating multiple chain-of-thought paths and submitting all of them to the LLM-as-a-Judge module (analogous to the fourth row in Table~\ref{tab:ablation}).
Although this approach naturally improves accuracy by exploring more reasoning paths, it also substantially increases computational cost.
As shown in Table \ref{tab:cost_analysis}, \name{} attains higher accuracy (78.8\%) at a lower total cost (\$0.140), even with greater input token usage, indicating that its extraction and merging steps effectively focus on task-relevant information and minimize redundant reasoning.
Since end-to-end latency may vary across hardware and network configurations, we report token consumption as a consistent measure of inference efficiency.\looseness=-1
\begin{table}[t]
\centering
\small
\caption{\textbf{Generalization to reasoning LLMs.} Average test accuracy on M3-Exam using Qwen3-8B with and without think-mode. The best and second best scores are highlighted in \textbf{bold} and \underline{underline}, respectively.}
\begin{tabular}{lccc}
\toprule
Methods & All & High & Low \\
\midrule
\multicolumn{4}{l}{\textbf{w/o Think}} \\
\textsc{Native-Basic} & 37.8 & 44.6 & 31.0 \\
\textsc{Eng-Basic} & 45.6 & 49.2 & 42.0 \\
\textsc{\name{} (Ours)} & \textbf{67.3} & \textbf{80.0} & \textbf{54.7} \\
\midrule
\multicolumn{4}{l}{\textbf{w/ Think}} \\
\textsc{Native-Basic} & 49.5 & 67.1 & 31.8 \\
\textsc{Eng-Basic} & 45.6 & 49.2 & 42.0 \\
\textsc{Trans-Google} & 49.3 & 65.4 & 33.1 \\
\textsc{\name{} (Ours)} & \underline{65.0} & \underline{79.4} & \underline{50.7} \\
\bottomrule
\end{tabular}
\label{tab:think_experiment}
\vspace{-0.2in}
\end{table}

\paragraph{Additional experiment with thinking models.}
We further evaluate our framework using the recent Qwen3-8B model \citep{qwen_docs}, which provides an optional "think-mode".
We conduct experiments on the full M3-Exam dataset, comparing multiple baselines and our \name{} pipeline under both think-mode settings.
As shown in Table~\ref{tab:think_experiment}, enabling the think-mode substantially benefits native prompting (from 37.8\% to 49.5\%), suggesting that internal reasoning processes can be helpful in some scenarios.
However, this improvement is not consistent across settings, as \textsc{Eng-Basic} exhibits identical performance (45.6\%) regardless of whether think-mode is enabled, indicating the limited generalizability of think-mode under prompt-based reasoning.
In contrast, \name{} consistently enhances accuracy in both configurations (67.3\% without think-mode and 65.0\% with think-mode).
This shows that the extraction component, rather than implicit reasoning augmentation of think-mode, is the main contributor to \name{}'s effectiveness.
\paragraph{Qualitative examples.} 
As shown in Figure \ref{fig:examples}, we provide a qualitative example to support further analysis.
(1) The original question is written in Javanese, and (2) the translated version indicates that it is asking about the specific name for the enclosure in which “Pagupon” is kept. 
(3) The response generated by the Eng-CoT method involves interpreting each option by explaining its meaning in English. 
It then concludes that the question is related to domesticated animals, the incorrect selection of option A. 
Although one might superficially associate “cage” with “chicken,” this interpretation is culturally and contextually incorrect in Javanese, where the term “Pagupon” is mainly used for pigeons or doves. 
(4) In contrast, our task-relevant knowledge extraction process correctly identifies D as the answer. 
(5) Finally, the LLM-as-a-Judge confirms that the answer is D instead of A.

\paragraph{Failure mode: irrelevant contextualization.}
While the extraction step is designed to provide helpful cultural and contextual grounding, it can sometimes introduce irrelevant context when the query concerns a globally identifiable entity rather than culture-specific knowledge.
As shown in Figure \ref{fig:mkqa_fail_case}, the query asks, “Whose song is ‘Wake Me Up Before You Go-Go’?” 
While the correct answer is “Wham!”, the extracted explanation regarding Japanese cultural knowledge incorrectly identifies the song as performed by the Japanese artist Koda Kumi, introducing irrelevant local cultural information. 
This mismatch illustrates that the extraction step can misinterpret the query's intent, especially when the query refers to a globally recognized entity rather than a culturally specific one.

\begin{figure}[t]
    \small
    \centering
    \begin{tabular}{p{0.9\linewidth}}
    \toprule 
    \textbf{(1) Native query:}  
    \begin{CJK}{UTF8}{min} ウキウキ・ウェイク・ミー・アップは誰の歌ですか。 
    \end{CJK} (Whose song is ‘Wake Me Up Before You Go-Go’?)
\\
\midrule 
\textbf{(2) Correct Answer:}  \begin{CJK}{UTF8}{min}ワム! \end{CJK} (Wham!)
\\
\midrule 
\textbf{(3) Extracted information:} The song \begin{CJK}{UTF8}{min}"ウキウキ・ウェイク・ミー・アップ" \end{CJK}(Ukiuki Wake Me Up) is performed by the Japanese artist Koda Kumi...She is a prominent figure in the J-Pop scene ...
\\ \bottomrule
    \end{tabular}
    \vspace{-0.05in}
    \caption{\textbf{Failure mode on MKQA.} The extraction step injects irrelevant language-specific context into a query whose correct answer is Wham!, illustrating how unnecessary contextualization can mislead the model.}
    \label{fig:mkqa_fail_case}
    \vspace{-0.1in}
\end{figure}

\section{Related Works} 

\noindent\textbf{Multilingual prompting of LLMs.} 
Recent advances in multilingual prompting have shown that translation-based strategies can substantially improve LLM performance on non-English tasks.
XLT \citep{huang2023not} proposes cross-lingual thought prompting, which translates non-English inputs into English for step-by-step reasoning, yielding strong multilingual performance.
Similarly, \citet{liu2024translation} shows that simple translation pipelines can yield competitive outcomes, although they may lose language-specific nuances.
Despite these gains, translation-based approaches often fail to capture language-specific nuances, whereas our framework instead directly synthesizes task-relevant contextual signals from the model.

\paragraph{Synthetic data and context extraction using LLMs.}
LLMs have recently been explored as sources of synthetic supervision, where models generate their own auxiliary data or contextual signals to enhance downstream performance.
Recitation \citep{sun2023recitation} introduces a self-contained prompting strategy that allows LLMs to generate task-aligned knowledge without an external retrieval.
Query2Doc \citep{wang2023query2doc} expands queries into pseudo-documents, improving information grounding for open-domain and QA tasks.
Other works explore self-generated supervision or adaptive retrieval using model-generated signals \citep{wang2023self, li2024self}.

\paragraph{Response merging and self-evaluation with LLMs.}
Recent research increasingly explores ways to integrate multiple LLM-generated responses through structured synthesis or self-assessment.
Multi-agent debate approaches \citep{duimproving, khandebating} show that iterative discussions among models can enhance reasoning quality and reduce factual errors.
Likewise, \citet{liang2023encouraging} proposes balancing divergent and convergent thinking, where LLMs first generate diverse hypotheses and then consolidate them into a coherent final answer.
Building on this idea, we adopt the LLM-as-a-Judge paradigm to assess and combine extracted synthetic contexts with reasoning-focused outputs, ensuring that the final response remains both logically consistent and semantically faithful to the intended objective.


\section{Conclusion}
In this paper, we propose \name{}, a multilingual prompting framework designed to enhance the handling of non-English queries in LLMs.
We address the limitations of previous approaches by (1) explicitly extracting synthetic context from LLMs and (2) integrating it with reasoning pathways through an LLM-as-a-Judge mechanism to produce contextually appropriate and logically coherent responses.
Empirical results show the effectiveness of \name{}, particularly in low-resource language settings.
These highlight the importance of incorporating query-relevant knowledge for non-English inputs and underscore the need for continued research in multilingual prompting from this view.

\newpage
\section*{Limitations} 
While our approach yields notable performance improvements, the computational cost associated with multiple LLM inferences remains a concern. 
Nevertheless, as shown in Table~\ref{tab:cost_analysis}, the performance gains are substantial enough to justify the increased cost. 
Notably, the \textsc{Eng-CoT} method performs three rounds of distinct reasoning followed by a merging step, resulting in a total of three inferences—comparable to our framework. 
However, despite incurring a higher cost, \textsc{Eng-CoT} achieves lower accuracy than our method, indicating that simply increasing the number of inference rounds does not necessarily lead to better results.

Also, integrating retrieval-augmented generation (RAG) to provide external cultural or contextual references could mitigate this limitation by supplementing the model with relevant knowledge.
Nonetheless, even when compared against RAG-based baselines (\textsc{RAG (Native)} and \textsc{RAG (Eng)}), \name{} consistently outperforms them on most benchmarks, demonstrating that its internally extracted context is both more targeted and efficient than externally retrieved passages.
Furthermore, our method is also effective for improving the multilingual Aya model (see Appendix~\ref{sec:aya}), which is explicitly trained to handle low-resource languages, confirming that \name{} generalizes effectively without additional training or external retrieval resources.

\section*{Broader Impact and Ethical Implications}
We believe that our work can help make LLM advancements more accessible to non-English users. 
As LLMs become increasingly integral to digital communication, it is important to ensure fair access for people who speak different languages.
By reducing language barriers, our research takes a key step toward making AI technologies less biased and more inclusive.

\section*{Acknowledgments}

All authors are affiliated with the Department of Artificial Intelligence at Yonsei University.
This research was supported in part by Institute for Information \& communications Technology Planning \& Evaluation (IITP) grant funded by the Korea government (MSIT) (No. RS-2020-II201361, Artificial Intelligence Graduate School Program (Yonsei University); No. RS-2025-02215344, Development of AI Technology with Robust and Flexible Resilience Against Risk Factors; No. RS-2025-25442405, Development of a Self-Learning World Model-Based
AGI System for Hyperspectral Imaging).

\bibliography{custom}
\appendix
\newpage

\section{Experimental Details}
This section provides comprehensive details on the experimental setup, including dataset descriptions, baseline methods, implementation specifics, and evaluation aspects used in our NLP experiments.

\subsection{Datasets}\label{supp:dataset}
The following tables list example instances from the M3-Exam (Table \ref{tab:dataset_m3}), MKQA (Table \ref{tab:dataset_mkqa}), XNLI (Table \ref{tab:dataset_xnli}), XCOPA (Table \ref{tab:dataset_xcopa}), and GlobalOpinionQA (Table \ref{tab:dataset_global}) datasets. The basic prompts used for these datasets are presented in Table \ref{tab:basic-prompt}.

\paragraph{M3-Exam}
M3-Exam \citep{zhang2023m3exam} is a multilingual multiple-choice question dataset designed to evaluate LLMs across eight languages: Afrikaans, Chinese, Italian, Javanese, Portuguese, Swahili, Thai, and Vietnamese. It consists of real-world exam questions collected from different countries, ensuring a diverse and authentic assessment of model performance. The dataset includes questions from South Africa for Afrikaans, China for Chinese, Italy for Italian, Indonesia for Javanese, Brazil for Portuguese, Kenya for Swahili, Thailand for Thai, and Vietnam for Vietnamese. Each language's dataset is further categorized into up to four subject areas, reflecting the structure of standardized exams in various educational systems. By incorporating multiple languages and educational contexts, M3-Exam offers a comprehensive benchmark for evaluating multilingual LLMs, particularly in underrepresented languages. Additionally, only text-based questions were used, explicitly excluding any questions requiring image-based inputs. Furthermore, questions that rely heavily on external background knowledge for answering were removed to ensure a fair evaluation of LLM capabilities based solely on textual reasoning. As a result, a total of 5,857 questions were selected for evaluation, providing a focused and controlled benchmark for assessing multilingual model performance, particularly in underrepresented languages.\footnote{The M3-Exam dataset was obtained from the official repository: \url{https://github.com/DAMO-NLP-SG/M3Exam}.} 

\begin{table*}[]
\centering
\small
\caption{Example data from M3-Exam.}
\vspace{-0.1in}
\renewcommand{\arraystretch}{1.2} 
\begin{tabular}{lp{11cm}} 
\toprule
\textbf{Component} & \textbf{Content} \\ 
\midrule
Question & Una scuola ha dieci classi, con una media di 22 alunni per classe. 
Le classi con 21 alunni sono sei; le classi con 24 alunni sono tre. 
Quanti alunni ci sono nella decima classe? \\
\midrule
Options & A. 20 \quad B. 22 \quad C. 23 \quad D. 25 \\
\midrule
Ground Truth & B \\ 
\bottomrule

\end{tabular}
\vspace{-0.05in}
\label{tab:dataset_m3}
\vspace{-0.1in}
\end{table*}

\paragraph{MKQA}  
MKQA \citep{longpre-etal-2021-mkqa} (Multilingual Knowledge Questions and Answers) is a dataset designed for evaluating open-domain question-answering capabilities across multiple languages. 
It is derived from the Natural Questions (NQ) dataset, which originally contains English questions and corresponding answers. 
For this study, we focused on nine languages: German, Spanish, French, Japanese, Russian, Thai, Turkish, Vietnamese, and Chinese. 
We selected 100 questions per language, resulting in a total of 900 samples for evaluation. 
Additionally, we excluded data with an entity type labeled as `long-answer' or `not answerable.'
The model’s performance was measured using the F1 score, ensuring a robust assessment of multilingual generation capabilities.\footnote{The MKQA dataset was obtained from the official repository: \url{https://github.com/apple/ml-mkqa}.}

\begin{table*}[]
\centering
\small
\vspace{0.1in}
\caption{Example data from MKQA.}
\renewcommand{\arraystretch}{1.2} 
\begin{tabular}{lp{11cm}} 
\toprule
\textbf{Component} & \textbf{Content} \\ 
\midrule
Question & en moyenne quelle longueur a le wagon de train de marchandises\\
\midrule
Answer Type & number\_with\_unit \\ 
\midrule
Ground Truth & 55.0 67.0 pied \\ 
\midrule
aliases & 55 67 pied \\ 
\bottomrule
\end{tabular}
\vspace{-0.05in}
\label{tab:dataset_mkqa}
\vspace{-0.1in}
\end{table*}

\paragraph{XNLI}
XNLI \citep{conneau-etal-2018-xnli} (Cross-lingual Natural Language Inference) is a multilingual dataset designed to evaluate a model's ability to understand and reason across different languages. 
It is derived from the MultiNLI dataset, which consists of English sentence pairs labeled for natural language inference (NLI) tasks—entailment, contradiction, or neutral. 
For this study, we conducted evaluations in 14 languages: French, Spanish, German, Greek, Bulgarian, Russian, Turkish, Arabic, Vietnamese, Thai, Chinese, Hindi, Swahili, and Urdu. 
We selected 100 samples per language, resulting in a total of 1400 samples for evaluation. 
Model performance was measured using accuracy, ensuring a reliable assessment of cross-lingual natural language understanding.\footnote{The XNLI dataset was obtained from Hugging Face: \url{https://huggingface.co/datasets/facebook/xnli}.}
\begin{table*}[h]
\centering
\small
\vspace{0.1in}
\caption{Example data from XCOPA.}
\vspace{-0.05in}
\renewcommand{\arraystretch}{1.2} 
\begin{tabular}{lp{11cm}} 
\toprule
\textbf{Component} & \textbf{Content} \\ 
\midrule
Question & effect \\
\midrule
Premise & Avevo svuotato le mie tasche. \\
\midrule
Choice 1 &  Estrassi la matrice di un biglietto.\\ 
\midrule
Choice 2 &  Trovai un'arma.\\ 
\midrule
Label &  0 (Choice 1)\\ 
\bottomrule
\end{tabular}
\label{tab:dataset_xcopa}
\vspace{-0.1in}
\end{table*}

\paragraph{XCOPA}
XCOPA \citep{ponti-etal-2020-xcopa} (Cross-lingual Choice of Plausible Alternatives) is a multilingual dataset designed to evaluate a model's ability to perform causal reasoning across different languages. 
It is an extension of the original COPA dataset, which consists of sentence pairs where one must determine the more plausible cause or effect. 
For this study, we conducted evaluations in 11 languages: Estonian, Haitian Creole, Indonesian, Italian, Quechua, Swahili, Tamil, Thai, Turkish, Vietnamese, and Chinese. 
We selected 100 samples per language, resulting in a total of 1100 samples for evaluation. Model performance was measured using accuracy, providing a standardized assessment of cross-lingual causal reasoning abilities.\footnote{The XCOPA dataset was obtained from Hugging Face: \url{https://huggingface.co/datasets/cambridgeltl/xcopa}.}

\begin{table*}[h]
\centering
\small
\vspace{0.1in}
\caption{Example data from XNLI.}
\renewcommand{\arraystretch}{1.2} 
\begin{tabular}{lp{11cm}} 
\toprule
\textbf{Component} & \textbf{Content} \\ 
\midrule
Premise & Nun, daran dachte ich nicht einmal, aber ich war so frustriert, dass ich am Ende doch mit ihm redete.\\
\midrule
Hypothesis & Wir hatten ein tolles Gespräch. \\
\midrule
Label & 1 (neutral) \\ 
\bottomrule
\end{tabular}
\vspace{-0.05in}
\label{tab:dataset_xnli}
\end{table*}

\begin{table*}[ht]
\centering
\small
\vspace{0.1in}
\caption{Example data from GlobalOpinionQA.}
\renewcommand{\arraystretch}{1.2} 
\begin{tabular}{lp{11cm}} 
\toprule
\textbf{Component} & \textbf{Content} \\ 
\midrule
Question & Glauben Sie, dass die Regierung Saudi-Arabiens die persönlichen Freiheiten ihrer Bürger respektiert, oder glauben Sie das nicht? \\
\midrule
Options & (A) Ja, respektiert persönliche Freiheiten \quad (B) Nein, respektiert persönliche Freiheiten nicht \\
\midrule
Ground Truth & B \\ 
\bottomrule
\end{tabular}
\vspace{-0.05in}
\label{tab:dataset_global}
\vspace{-0.1in}
\end{table*}

\paragraph{GlobalOpinionQA}  
GlobalOpinionQA \citep{durmus2024towards} is a multilingual question-answering dataset designed to evaluate LLMs on opinion-based questions collected from diverse countries. 
The dataset consists of survey-based multiple-choice questions covering a range of global topics. 
For this study, we conducted evaluations using data from 34 countries, including Greece, Sweden, China (non-national sample), Tunisia, Malaysia, Vietnam, Argentina, Russia, Egypt, Indonesia, Jordan, Mexico, Pakistan, Palestinian territories, Turkey, Ukraine, Kenya, France, Germany, Lebanon, Peru, Poland, South Korea, Italy, Spain, Brazil, Chile, Japan, Venezuela, Senegal, the Netherlands, Uganda, the Philippines, and Ethiopia.  
To ensure a reliable evaluation, we only included languages with at least 40 samples in the dataset. 
Since the dataset originally included the answer distribution from multiple respondents within the same country, we used the dataset from \citet{kim2024few}, which converts these distributions into a single-answer format by selecting the choice with the highest probability. 
This standardized format enables a more consistent assessment of model performance across different linguistic and cultural contexts.
Model performance was measured using accuracy, providing insights into the model’s ability to align with human responses across various regions.  


\begin{table*}[b] 
\vspace{-0.1in}
\caption{Test accuracy of three different LLMs on M3-Exam. The best and second best scores are highlighted in \textbf{bold} and \underline{underline}, respectively.}
\vspace{-0.1in}
\begin{center}
\resizebox{0.9\textwidth}{!}{
\begin{tabular}{l|ccc|ccc|ccc} 
\toprule
\textbf{Models ($\rightarrow$)} & \multicolumn{3}{c}{\textbf{GPT-4o}} & \multicolumn{3}{c}{\textbf{Claude-Haiku}} & \multicolumn{3}{c}{\textbf{Llama-3.1-8B}} \\
\cmidrule(lr){2-4} \cmidrule(lr){5-7} \cmidrule(lr){8-10}
\textbf{Methods ($\downarrow$)} & All & High & Low & All & High & Low & All & High & Low \\ 
\midrule
\textsc{Native-Basic} & 78.1 & 84.5 & 71.7 & 67.4 & 76.4 & 58.3 & 49.8 & 60.9 & 38.7 \\
\textsc{Trans-Google} & 68.4 & 78.3 & 58.5 & 60.1 & 67.5 & 52.7 & \underline{53.9} & \underline{64.0} & \underline{43.8} \\
\textsc{XLT}          & 81.8 & 89.2 & 74.3 & 74.0 & 84.4 & 63.6 & 25.8 & 30.7 & 20.8 \\
\textsc{Native-Cot}   & \underline{83.9} & \underline{90.1} & 77.7 & 74.8 & \underline{84.8} & 64.8 & 36.8 & 46.6 & 27.1 \\
\textsc{Eng-Cot}      & 82.9 & 88.1 & \underline{76.7} & \underline{75.4} & 83.9 & \textbf{66.8} & 40.4 & 47.2 & 33.6 \\ 
\midrule 
\textbf{\name{} (Ours)} & \textbf{85.7} & \textbf{91.5} & \textbf{79.8} & \textbf{75.6} & \textbf{85.8} & \underline{65.4} & \textbf{56.9} & \textbf{69.3} & \textbf{44.5} \\ 
\bottomrule
\end{tabular}%
}
\end{center}
\vspace{-0.1in}
\label{tab:diff_llm_full}
\end{table*}
\begin{table*}[]
\caption{Basic prompts used for different benchmarks.}
\centering
\small
\renewcommand{\arraystretch}{1.2} 
\begin{tabular}{lp{11cm}} 
\toprule
\textbf{Benchmark} & \textbf{Basic Prompt} \\ 
\midrule
M3-Exam &
  The following is a multiple-choice question. \newline
  \{Question\} \newline
  \{Options\}\newline
  You should provide the final answer at the end in the format: `Answer: {[}option{]}'  \\
\midrule
MKQA &
  Answer the question in one or a few words in English:\newline
  \{Question\} \newline
  You should provide the final answer at the end in the format: `Answer: ' \\
\midrule
XNLI &
  \{Premise\} \newline
  Based on the previous passage, is it true that \{Hypothesis\}? \newline
  Answer should be in the format of "Answer: {[}Yes or No or Maybe{]}".  \\
\midrule
XCOPA &
  Here is a premise: \{Premise\}.\newline
  What is the \{Question\}? \newline
  Help me pick the more plausible option:\newline-choice1: \{Choice1\}, -choice2: \{Choice2\} \newline
  You should only choose one option for your answer.\newline
  You should answer the question in the format of `Answer: {[}1 or 2{]}'  \\
\bottomrule
\end{tabular}
\label{tab:basic-prompt}
\end{table*}

\subsection{Measurement details}\label{supp:measurement}
To distinguish high- and low-resource languages for each dataset, we used performance measured on native queries. 
Since the classification relies on the model's performance, the high/low-resource split may vary across different models. 
Table \ref{tab:high_low_lang} shows the split for M3-Exam, MKQA, XNLI, XCOPA, and GlobalOpinionQA based on GPT-4o-mini.

\begin{table*}[t]
\centering
\small
\caption{High- and low-resource subsets for each dataset. For GlobalOpinionQA, we report countries.}
\renewcommand{\arraystretch}{1.2}
\begin{tabular}{lp{5cm}p{5cm}} 
\toprule
\textbf{Dataset} & \textbf{High-resource languages} & \textbf{Low-resource languages} \\ 
\midrule
M3-Exam & Afrikaans, Chinese, Portuguese, Vietnamese & Italian, Javanese, Swahili, Thai \\
\midrule
MKQA & German, Spanish, French, Japanese, Thai & Russian, Turkish, Vietnamese, Chinese \\
\midrule
XNLI & French, German, Greek, Bulgarian, Russian, Thai, Hindi & Spanish, Turkish, Arabic, Vietnamese, Chinese, Swahili, Urdu \\ 
\midrule
XCOPA & Estonian, Indonesian, Italian, Thai, Turkish, Vietnamese & Haitian Creole, Quechua, Swahili, Tamil, Chinese \\
\midrule
GlobalOpinionQA & Greece, Sweden, China (Non-national), Ukraine, Kenya, France, Germany, Lebanon, Poland, S.Korea, Italy, Spain, Brazil, Chile, Japan, Senegal, Netherlands & Tunisia, Malaysia, Vietnam, Argentina, Russia, Egypt, Indonesia, Jordan, Mexico, Pakistan, Palest. ter, Turkey, Peru, Venezuela, Uganda, Philippines, Ethiopia \\
\bottomrule
\end{tabular}
\label{tab:high_low_lang}
\end{table*}

\begin{table*}[t]
\centering
\small
\caption{Prompts used for different baselines.}
\begin{tabular}{lp{10cm}} 
\toprule
\textbf{Settings} &
  \textbf{Content} \\ \midrule
Original Question &
  Pagupon Iku Jenenge Kandhang ....  \newline A. Pitik \quad B. Macan \quad C. Jaran \quad D. Dara \\ \midrule
Translated Question &
  Pagupon Is The Name Of The Cage .... \newline A. Chicken \quad B. Tiger \quad C. Horse \quad D. Dara \\
  \midrule
\textsc{Native-Basic} &
  Ing ngisor iki minangka pitakon pilihan ganda.\newline
  \{Original Question\}\newline 
  Sampeyan kudu nyedhiyakake jawaban pungkasan ing pungkasan nganggo format: `Wangsulan: {[}pilihan{]}'. \\ \midrule
\textsc{Eng-Basic} &
  The following is a multiple-choice question.\newline 
  \{Original Question\}\newline 
  You should provide the final answer at the end in the format: `Answer: [option]'.\\ 
  \midrule
\textsc{Trans-Google} &
  The following is a multiple-choice question.\newline \{Translated Question\}\newline You should provide the final answer at the end in the format: `Answer: {[}option{]}'. \\ \midrule
\textsc{XLT} &
  I want you to act as an question answering expert for Javanese.\newline
  Request: \{Original Question\}\newline
  You should retell the request in English.\newline
  You should do step-by-step answer to obtain an option.\newline
  You should step-by-step answer the request.\newline 
  You should tell me the answer in this format `Answer : {[}option{]}'. \\ \midrule 
\textsc{Native-Cot} &
 Ing ngisor iki minangka pitakon pilihan ganda.\newline 
 \{Original Question\}\newline 
 Ayo mikir kanthi bertahap nganggo basa Jawa, lan weneh jawaban pungkasan ing pungkasan kanthi format: `Jawaban: [pilihan]' \\ \midrule
\textsc{Eng-CoT} &
  The following is a multiple-choice question.\newline
  \{Original Question\}\newline
  Let's think step-by-step in English, and provide the final answer at the end in the format: `Answer: [option]' \\ 
\bottomrule
\end{tabular}
\label{tab:prompt_template}
\end{table*}

\subsection{Implementation details}\label{supp:details}
\paragraph{\name{} Details.} Table \ref{tab:prompt_template} below presents the prompts used for the baselines in our experiments.
When selecting few-shot examples, we ensured that they did not overlap with the test dataset. 
If a train or development dataset was available, we sampled examples from those datasets. 
In cases where only a test dataset was available, we excluded the few-shot examples from the evaluation set.
We used GPT-4o to generate answers for the few-shot examples, providing the ground-truth labels during selection to minimize errors and ensure their reliability.
For M3-Exam, which consists of four subject categories, we used four few-shot examples, ensuring representation across different subject categories as much as possible. 
For other datasets that do not have predefined categories, we used three examples per dataset.
In all cases, the examples were randomly selected to avoid selection bias. The few-shot examples are presented in Table \ref{tab:few_shot_m3}.

\paragraph{RAG Details.}
For the retrieval-augmented baselines, we used the Google Custom Search API\footnote{\url{https://developers.google.com/custom-search/}}
 to retrieve relevant web documents.
To ensure fairness with our experimental setup, we only utilized the top-1 ranked document for each query and truncated it to the first 100 words.
For \textsc{RAG (Native)}, we combined the retrieved document with native-language instructions, whereas \textsc{RAG (Eng)} used equivalent English instructions while keeping the query in the original language.
This setup allows us to evaluate the influence of instruction language on retrieval-augmented prompting while maintaining consistent retrieval conditions across both variants.
\begin{table*}[t]
\centering
\small
\caption{Few-shot examples used in the M3-Exam Afrikaans experiment.}
\renewcommand{\arraystretch}{1.2} 
\begin{tabular}{lp{11cm}} 
\toprule
\textbf{Component} & \textbf{Content} \\ 
\midrule
Question & Identifiseer die hoofsin in die onderstaande sin.\newline Gesonde mense drink elke dag in die winter water. \\
\midrule
Options & (A) Mense drink water.\quad(B) Mense drink elke dag water.\quad(C) Mense drink in die winter water. (D) Mense drink.
\\
\midrule
Category & Language \\ 
\midrule
Explanation & In the given sentence, \"Gesonde mense drink elke dag in die winter water,\" we are looking for the main clause, which is the complete thought or statement that provides the essential information without additional modifiers. The main subject is \"mense\" (people) and the main verb is \"drink\" (drink). Option (A) \"Mense drink water\" captures this essential information, as it includes the subject \"Mense\" and the verb \"drink,\" along with the object \"water,\" forming the core of the sentence without the additional details about when or how often the action occurs. Options B, C, and D include modifiers or less complete versions of the main statement. \\ 
\midrule
\midrule
Question & IOpwekking van elektrisiteit deur vallende water, word ... genoem. \\
\midrule
Options & (A) kernkrag \quad (B) hidro-elektrisiteit\quad(C) transformasie\quad(D) aardlekkasie
\\
\midrule
Category & Natural Science \\ 
\midrule
Explanation & The generation of electricity by falling water is termed \"hydro-electricity,\" which is option (B). This process involves converting the kinetic energy of falling or flowing water into electrical energy using turbines and generators. Unlike nuclear power (option A) which relies on nuclear reactions, or transformation (option C) which refers to changing one form of energy to another, hydro-electricity specifically pertains to water. Aardlekkasie (option D) refers to ground leakage, which is unrelated to the generation of electricity through water. Understanding the specific terminologies associated with different energy sources is crucial in selecting the correct answer. \\ 
\midrule
\midrule
Question & Om te verhoed dat die gruweldade van WWII weer plaasvind, het die `Algemene Vergadering van die VN' ........ in Desember 1948 aanvaar. \\
\midrule
Options & (A) die Wet op Afsonderlike Geriewe\quad(B) die paswette \quad(C) die Universale Verklaring van Menseregte\quad(D) die Suid-Afrkaanse Handves van Regte
\\
\midrule
Category & Social Science \\ 
\midrule
Explanation & The correct answer is (C) the 'Universal Declaration of Human Rights.' After the atrocities of World War II, the United Nations sought to ensure such events would not happen again by establishing a set of fundamental human rights that all nations should uphold. In December 1948, the General Assembly of the United Nations adopted the Universal Declaration of Human Rights, which outlines basic rights and freedoms to which all humans are entitled. Options (A), (B), and (D) are incorrect as they pertain to specific laws or documents related to South Africa and are unrelated to the global context addressed by the UN in 1948. \\ 
\midrule
\midrule
Question & Watter van die volgende getalle is 'n $\langle$b$\rangle$priemgetal$\langle$/b$\rangle$? \\
\midrule
Options & (A) 1\quad(B) 2\quad(C) 0\quad(D) 4
\\
\midrule
Category & Math \\ 
\midrule
Explanation & A prime number is defined as a natural number greater than 1 that has no positive divisors other than 1 and itself. In this context, understanding prime numbers is crucial. Option A (1) is not a prime number because it only has one divisor, which is itself. Option C (0) is not a prime number because it is not greater than 1. Option D (4) is not a prime number because it has divisors other than 1 and itself, namely 2. Therefore, the only option that fits the definition of a prime number is B (2), as it is greater than 1 and can only be divided by 1 and 2. \\ 
\bottomrule
\end{tabular}

\label{tab:few_shot_m3}
\end{table*}

\section{Additional Analyses}\label{supp:analyses}
\subsection{Full experiment on other datasets}
\begin{figure*}[t]
\centering
\includegraphics[width=0.8\textwidth]{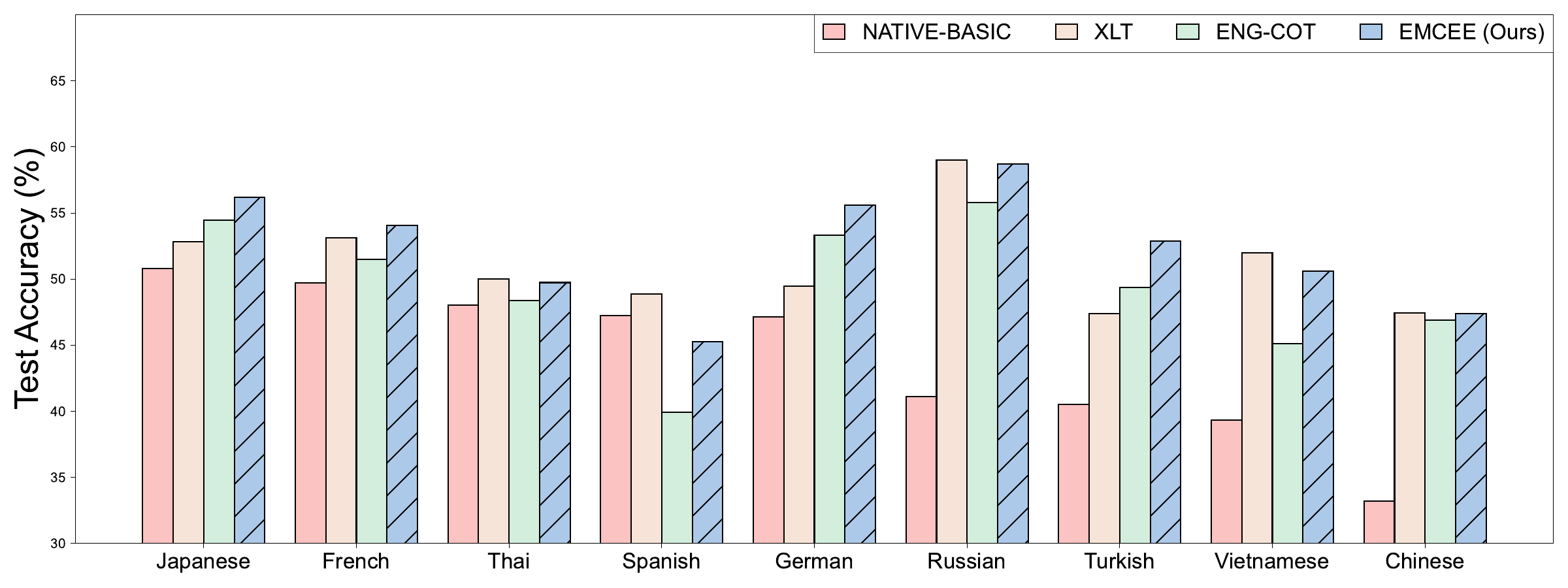}
\caption{Overall language-wise improvement for MKQA.}
\label{fig:graph_mkqa}
\end{figure*}

\begin{figure*}[t]
\centering
\includegraphics[width=0.8\textwidth]{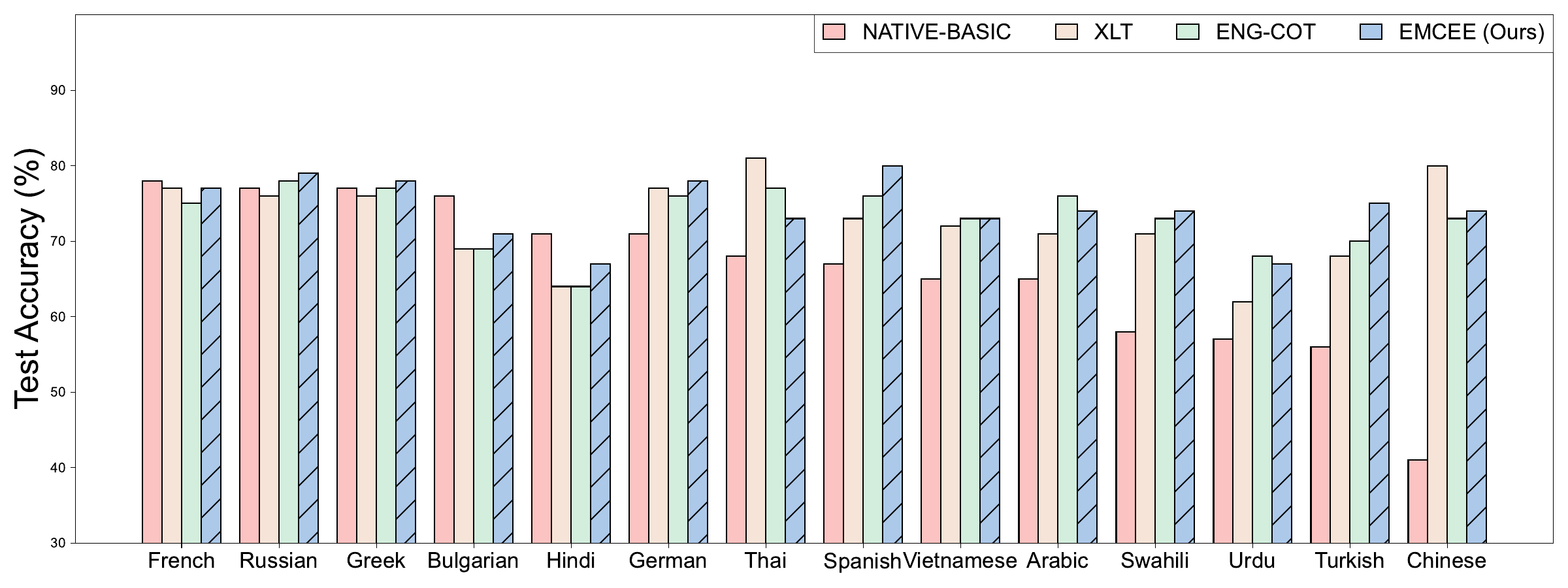}
\caption{Overall language-wise improvement for XNLI.}
\label{fig:graph_xnli}
\end{figure*}

\begin{figure*}[t]
\centering
\includegraphics[width=0.8\textwidth]{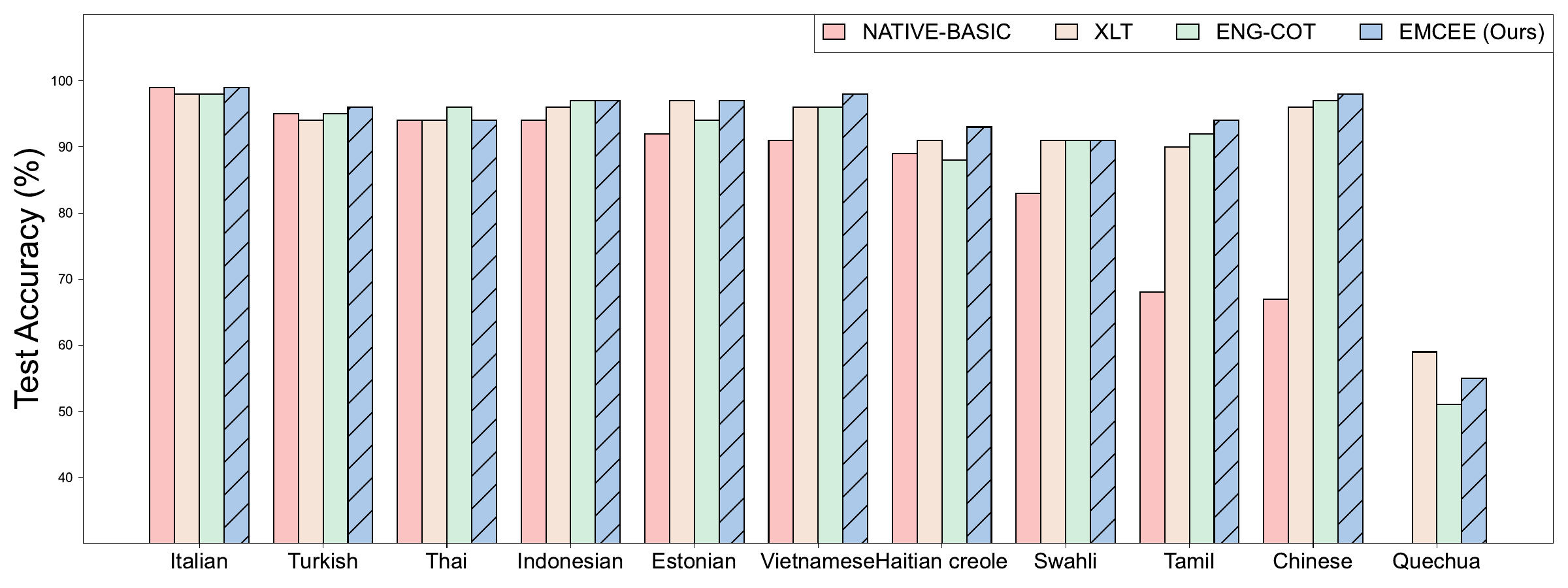}
\caption{Overall language-wise improvement for XCOPA.}
\label{fig:graph_xcopa}
\end{figure*}

The experimental results from Table \ref{tab:main} in the main paper are presented in graphical form. Specifically, the results for MKQA are shown in Figure \ref{fig:graph_mkqa}, XNLI in Figure \ref{fig:graph_xnli}, and XCOPA in Figure \ref{fig:graph_xcopa} to provide a clearer visualization of the findings.

\subsection{Effect of language information.}
We investigate the role of language information in \name{} by conducting an ablation study that progressively removes linguistic cues from the Extracting and Merging stages (Table~\ref{tab:lang_info}). 
In the full \name{} setup, both stages use native-language signals in the instruction and the query (Figure~\ref{fig:prompt}).
We begin from the original \name{} configuration and first remove instruction-level language signals from the Merging stage, while keeping the Extracting stage unchanged. 
We then further remove query-level language signals in Merging by replacing the original query with its English-translated version. 
Next, we remove language information from the Extracting stage as well, while still retaining the query in Merging. 
Finally, we remove both instruction- and query-level language signals from Merging, yielding a setting in which both stages operate without explicit native-language information.
As shown in Table~\ref{tab:lang_info}, performance remains relatively stable for some high-resource languages. 
However, the overall results gradually decline as language information is removed, with the largest drop occurring when both instruction- and query-level signals are absent. 
These findings highlight the importance of explicit language guidance in both stages: instruction-level signals help maintain task alignment, while query-level signals preserve linguistic nuances that are particularly important for cross-lingual generalization, especially in low-resource settings.
\begin{table}[H]
\caption{Accuracy on M3-Exam when removing language signals from instruction and query in the Extracting (ExT) and Merging (MeR) stages.}
\label{tab:lang_info}
\begin{tabular}{lcccc}
\toprule
\textbf{Methods} & \textbf{ExT} & \textbf{MeR} & \textbf{All / High / Low} \\ \midrule
\name{}       & both         & both          & 77.4 / 83.3 / 71.5        \\ \midrule
                 & both         & query         & 76.3 / 83.3 / 69.3        \\
                 & query        & query         & 76.4 / 84.1 / 68.8        \\
                 & no           & query         & 75.5 / 83.5 / 67.5        \\
                 & no           & no            & 73.1 / 82.9 / 63.4        \\ \bottomrule
\end{tabular}
\end{table}

\subsection{Adaptive setting with M3-Exam}
We conduct additional experiments comparing \name{} pipeline with a adaptive contextual reasoning approach.
Specifically, while \name{} performs explicit extraction and merging of cultural and contextual knowledge before reasoning, the adaptive variant prompts the model to decide dynamically whether to generate such context.
We sample 100 queries for each of the eight languages in the M3-Exam dataset (800 queries in total) and use the \texttt{gpt-4o-mini} model for evaluation.
The adaptive variant employs the following system prompt, enabling the model to determine on its own whether to produce a contextual note:

\begin{quote}
\small
Whenever you decide that a cultural/contextual note is required, output exactly: \\
\texttt{Explanation: <your explanation here>} \\
Otherwise, output exactly: \\
\texttt{Explanation: None} \\
Do not include any other words, headings, or punctuation.
\end{quote}

\noindent We compare the following two configurations:
1) \textsc{Adaptive}: Standard Chain-of-Thought prompting augmented with the above adaptive instruction.
2) \name{}: Our original pipeline comprising extraction, reasoning, and merging stages.
As summarized in Table~\ref{tab:adaptive_experiment}, the adaptive approach achieves an average accuracy of 69.9\%, whereas \name{} attains 76.1\%.
These results indicate that explicitly extracting query-relevant knowledge for every query and subsequently determining its relevance yields stronger and more consistent improvements than relying on the model to decide on-the-fly.
\begin{table}[H]
\caption{Comparison between adaptive contextual reasoning (\textsc{Adaptive}) and \name{}.
The best results in each column are shown in \textbf{bold}.}
\centering
\small
\begin{tabular}{lccc}
\toprule
\textbf{Methods} & \textbf{All} & \textbf{High} & \textbf{Low} \\
\midrule
\textsc{Adaptive} & 69.9 & 79.5 & 60.3 \\
\name{} & \textbf{76.1} & \textbf{84.5} & \textbf{67.8} \\
\bottomrule
\end{tabular}
\label{tab:adaptive_experiment}
\end{table}

\subsection{Extraction in native language} \label{sec:ext_native}
To examine the impact of instruction language on our method, we conduct an additional ablation study comparing native language and English language instructions for the extraction stage. While our main experiments adopt English instructions, reflecting the fact that LLMs are predominantly trained and optimized on English data~\citep{meta2024llama3.1,liu2024translation}, we also test the effect of using native language instructions for synthetic knowledge extraction. Previous work~\citep{huang2023not} has shown that introducing English elements into multilingual prompts, for example for disfluency correction or inference tasks, often enhances model performance, which aligns with our design choice.

\noindent We evaluate 100 queries per language across eight languages in the M3-Exam (800 queries in total). In one condition, the model is prompted to extract knowledge using a native language instruction (\textsc{Native-Extract}). As shown in Table~\ref{tab:native_ext}, the native language extraction achieves an average accuracy of 70.8\%, while the English instruction variant slightly outperforms it with 72.5\%. These results suggest that English based instructions remain advantageous, likely due to the models' pretraining bias toward English, although native language prompting remains a promising direction for future research.
\begin{table}[H]
\centering
\small
\caption{Comparison between native language and English instructions for the extraction stage. The best results in each column are shown in \textbf{bold}.}
\begin{tabular}{lccc}
\toprule
\textbf{Methods} & \textbf{All} & \textbf{High} & \textbf{Low} \\
\midrule
\textsc{Native-Extract} & 70.8 & 78.8 & 62.8 \\
\textsc{Eng-Extract (Ours)} & \textbf{72.5} & \textbf{79.5} & \textbf{65.6} \\
\bottomrule
\end{tabular}
\label{tab:native_ext}
\end{table}

\subsection{Robustness across multiple runs}
To evaluate robustness with respect to prompt construction and few-shot exemplar selection, we repeated all methods over three independent random seeds, where each run used a different randomly sampled set of few-shot examples. Table~\ref{tab:multi_runs} reports the mean and standard deviation across the three runs.
Overall, the variance across seeds is consistently small for all methods. Native-Basic achieves $68.0 \pm 2.4$, Eng-CoT achieves $72.9 \pm 1.6$, XLT achieves $70.5 \pm 0.7$, and \name{} achieves $75.2 \pm 2.1$. 
Importantly, the performance gains of \name{} remain larger than the standard deviations of the baselines, and its improvement over Eng-CoT and XLT is preserved across all runs. These results indicate that the effectiveness of \name{} is not driven by a particular random seed or a specific selection of few-shot exemplars, but reflects a stable improvement across different prompt instantiations.
\begin{table}[H]
\centering
\small
\caption{Robustness across three independent runs. The best results in each column are shown in \textbf{bold}.}
\label{tab:multi_runs}
\begin{tabular}{lc}
\toprule
\textbf{Methods} & \textbf{Mean $\pm$ Std} \\
\midrule
Native-Basic & $68.0 \pm 2.4$ \\
Eng-CoT & $72.9 \pm 1.6$ \\
XLT & $70.5 \pm 0.7$ \\
\name{} \textsc{(Ours)} & $\mathbf{75.2 \pm 2.1}$ \\
\bottomrule
\end{tabular}
\end{table}

\subsection{Merging step with different combinations}
One possible concern is that the gain of \name{} may stem simply from the merging procedure rather than from the complementary roles of its two components. To examine this possibility, we evaluated several alternative combinations under the same merge-based framework on the full M3-Exam dataset with GPT-4o-mini. As shown in Table~\ref{tab:merge_combinations}, \name{}, which combines \textsc{Extract} and \textsc{Eng-CoT}, achieves the best overall performance among all tested combinations.
Specifically, alternative combinations such as \textsc{Native-Basic + Native-CoT} (73.1), \textsc{Eng-Basic + Eng-CoT} (74.4), \textsc{Native-Basic + Eng-CoT} (74.6), \textsc{Extract + RAG} (74.5), and \textsc{Eng-CoT + RAG} (73.2) all remain below \textsc{Extract + Eng-CoT (\name{})}, which reaches 77.4 average accuracy. This result suggests that the benefit does not arise from aggregation alone. Rather, the specific combination of what is being merged is crucial.
We hypothesize that the strength of \name{} comes from the complementary nature of its two paths. The \textsc{Extract} step provides multilingual, culturally grounded synthetic context by explicitly eliciting background knowledge that the model may fail to recall when directly answering non-English questions. 
In contrast, the \textsc{Eng-CoT} step offers a stable reasoning path by leveraging the model's generally stronger reasoning ability in English. The empirical results indicate that combining these two components is substantially more effective than other merge-based alternatives.

\begin{table}[H]
\centering
\small
\caption{Merge-based combinations on M3-Exam. The best results in each column are shown in \textbf{bold}.}
\label{tab:merge_combinations}
\resizebox{\columnwidth}{!}{%
\begin{tabular}{lccc}
\toprule
\textbf{Methods} & \textbf{All} & \textbf{High} & \textbf{Low} \\
\midrule
\textsc{Native-Basic + Native-CoT} & 73.1 & 81.5 & 64.7 \\
\textsc{Eng-Basic + Eng-CoT} & 74.4 & 83.4 & 65.4 \\
\textsc{Native-Basic + Eng-CoT} & 74.6 & \textbf{83.5} & 65.7 \\
\textsc{Extract + RAG} & 74.5 & 83.0 & 65.9 \\
\textsc{Eng-CoT + RAG} & 73.2 & 81.9 & 64.5 \\
\textsc{Extract + Eng-CoT (\name{})} & \textbf{77.4} & 83.3 & \textbf{71.5} \\
\bottomrule
\end{tabular}%
}
\end{table}
\subsection{Extended RAG baselines}
To better understand whether \name{} simply benefits from replacing retrieval with generation, we compare it against stronger retrieval-based baselines with multiple configurations. 
In addition to a basic single-passage setting, we evaluate full-passage retrieval, multi-passage retrieval, and a dense reranking pipeline. Table~\ref{tab:rag_baselines} summarizes the results over 100 queries for each M3-Exam language.
Among the RAG variants, the strongest baseline is the setting with five retrieved passages followed by reranking, which achieves 71.3 average accuracy. 
However, this result still falls substantially short of \name{}, which achieves 76.0 on the same evaluation subset. 
The gap is particularly notable in the low-resource setting, where the reranked RAG baseline reaches 62.0 while \name{} achieves 69.3.
These results suggest that the advantage of \name{} is not simply due to using more context or a more elaborate inference pipeline. 
Instead, explicitly eliciting the model's own multilingual and culturally grounded knowledge can be more effective than relying solely on externally retrieved passages, especially in multilingual scenarios where retrieval may be sparse, noisy, or culturally misaligned.
\begin{table}[H]
\centering
\small
\caption{Comparison of extended RAG baselines. Best results are shown in \textbf{bold}.}
\label{tab:rag_baselines}
\resizebox{\columnwidth}{!}{%
\begin{tabular}{lccc}
\toprule
\textbf{Methods} & \textbf{All} & \textbf{High} & \textbf{Low} \\
\midrule
\multicolumn{4}{l}{\textbf{RAG}} \\
One passage (100 words) & 70.6 & 79.8 & 61.5 \\
One passage (full text) & 70.4 & 80.0 & 60.8 \\
Five passages (100 words each) & 69.0 & 78.0 & 60.0 \\
Five passages + reranking & 71.3 & 80.5 & 62.0 \\
\midrule
\name{} \textsc{(Ours)} & \textbf{76.0} & \textbf{82.8} & \textbf{69.3} \\
\bottomrule
\end{tabular}%
}
\end{table}

\subsection{Comparison with a recitation-based baseline}
We further consider a recitation-based synthetic-context baseline inspired by prior work on self-generated contextual augmentation. 
In this setting, we run the \textsc{Extract} stage twice and merge the two generated outputs, yielding a baseline that uses only synthetic context generation without the reasoning path of \textsc{Eng-CoT}. 
We denote this variant as \textsc{Extract + Extract $\rightarrow$ Merge (RECITE)}.
As shown in Table~\ref{tab:additional_baseline}, this RECITE-style baseline achieves 72.4 average accuracy on the full M3-Exam dataset. 
Although it performs reasonably well, it remains below both \textsc{Eng-CoT + Eng-CoT $\rightarrow$ Merge} (74.6) and \textsc{Eng-CoT + Extract $\rightarrow$ Merge (\name{})} (77.4). 
This finding indicates that simply generating more synthetic text is not sufficient. 
Rather, the complementary combination of a knowledge-oriented path (\textsc{Extract}) and a reasoning-oriented path (\textsc{Eng-CoT}) is essential for achieving the full benefit of \name{}.

\begin{table}[H]
\centering
\small
\caption{Comparison with a recitation-based synthetic-context baselines. Best
results are shown in \textbf{bold}.}
\label{tab:additional_baseline}
\resizebox{\columnwidth}{!}{%
\begin{tabular}{lccc}
\toprule
\textbf{Methods} & \textbf{All} & \textbf{High} & \textbf{Low} \\
\midrule
\textsc{Eng-CoT + Eng-CoT} $\rightarrow$ \textsc{Merge} & 74.6 & 81.8 & 67.3 \\
\textsc{Extract + Extract} $\rightarrow$ \textsc{Merge (RECITE)} & 72.4 & 80.4 & 64.5 \\
\textsc{Eng-CoT + Extract $\rightarrow$ \textsc{Merge} (\name{})} & \textbf{77.4} & \textbf{83.3} & \textbf{71.5} \\
\bottomrule
\end{tabular}%
}
\end{table}

\section{Qualitative Examples}\label{supp:examples}
\subsection{Case study: when knowledge extraction can be detrimental}
In this section, we provide additional qualitative examples illustrating cases where the extraction step can negatively affect performance.
We consider two representative situations in which knowledge extraction may be counterproductive.
The first occurs when the generated explanation introduces context that is misaligned with the intent of the query, as illustrated by the failure case in the main text (Figure~\ref{fig:mkqa_fail_case}).
The second occurs when additional cultural or contextual grounding is unnecessary for solving the problem (Figure~\ref{fig:qualitative_4}).
In both cases, the reasoning-oriented Chain-of-Thought (CoT) response selected by the \name{} framework tends to provide a more reliable and efficient solution.
\begin{figure*}[ht]
    \centering
    \small
    \begin{tabular}{p{1.0\linewidth}}
    \toprule 
    \textbf{(1) Native query and Options:}  
    Wat is die waarde van 63,91 × 10? (What is the value of 63.91 × 10?) \\ (A) 6,391 (B) 63,910 (C) 6391 (D) 639,1
\\
\midrule 
\textbf{(2) Correct Answer:}  (D)
\\
\midrule 
\textbf{(3) Extracted information:} To find the value of 63.91 × 10, you need to multiply 63.91 by 10. When you multiply a number by 10, you simply shift the decimal point one place to the right.

\\ \bottomrule
    \end{tabular}
    \caption{Qualitative example from M3-Exam.}
    \label{fig:qualitative_4}
\end{figure*}

In Figure \ref{fig:qualitative_4}, the question simply asks for the result of a basic arithmetic operation ("What is the value of 63.91 × 10?"). 
Here, the extracted explanation redundantly elaborates on the concept of multiplying by ten, which adds unnecessary verbosity without contributing to correctness. 
Since contextual knowledge is irrelevant to mathematical reasoning, this case demonstrates how extraction may distract from the concise reasoning path provided by CoT.
Together, these examples highlight why the merging step in \textsc{\name{}}, which selects between extracted and reasoning-based responses is crucial. 
When the extracted context is misleading or irrelevant, the model can rely on CoT reasoning to produce a more accurate and faithful answer.

\subsection{More examples}
Additional qualitative examples with \name{} are shown in Figures \ref{fig:qualitative_1} and \ref{fig:qualitative_2}. 
All the examples are from M3-Exam with GPT-4o-mini.

\begin{figure*}[h]
    \centering
    \small
    \small
    \begin{tabular}{p{1.0\linewidth}}
    \toprule 
    \textbf{(1) Native query:}  
    Ni sentensi ipi sahihi? \newline
    A. Hewala! Nitakusaidia.
    B. Kefule! Naomba maji.
    C. Oyee! Tumeshindwa.
    D. Hamadi! Amekuja.
\\
\midrule 
\textbf{(2) Translated query:}
Which sentence is correct? \\
A. Hewala! I will help you.
B. Lock it! I'm asking for water.
C. Oye! We have failed.
D. Hamadi! He has come.
\\
\midrule 
\textbf{(3) Incorrect responses by \textsc{Eng-CoT}:} To determine which sentence is correct, we need to analyze each option based on standard Swahili grammar and usage.

A. "Hewala! Nitakusaidia." - "Hewala" is an informal greeting or expression, and "Nitakusaidia" means "I will help you." This sentence is grammatically correct.

B. "Kefule! Naomba maji." - "Kefule" is not a standard Swahili word or expression. "Naomba maji" means "I am asking for water," which is correct, but the first part makes the whole sentence questionable.

C. "Oyee! Tumeshindwa." - "Oyee" is an informal expression, and "Tumeshindwa" means "We have failed." This sentence is also grammatically correct.

D. "Hamadi! Amekuja." - "Hamadi" is a name, and "Amekuja" means "He/She has come." This sentence is correct as well.

Now, we have three sentences (A, C, and D) that are grammatically correct. However, the question asks for the most appropriate or commonly used expression. 

Considering the context of greetings and expressions, "Oyee!" is often used in celebratory or informal contexts, while "Hewala!" is also informal but less common. 

Thus, the most appropriate and commonly used expression in this context is option C.
\\
\midrule 
\textbf{(4) Correct response by {Extracting}:} 
To determine the correct sentence, we need to consider the context and usage of the words. In Swahili, "Hewala!" is an expression of excitement or encouragement, making option A a valid exclamation. However, "Kefule!" in option B is not a commonly recognized expression in Swahili, which makes it less likely to be correct. Option C, "Oyee! Tumeshindwa," uses "Oyee!" as an exclamation, but it conveys a negative context of defeat, which may not be appropriate depending on the intended meaning. Option D, "Hamadi! Amekuja," uses "Hamadi!" as a name or greeting, which is also acceptable. However, the most universally recognized and appropriate exclamation in a positive context is option A: "Hewala! Nitakusaidia," which expresses a willingness to help. Therefore, option A is the most suitable choice.
\\
\midrule 
\textbf{(5) Example of \name{} response comparing (3) \& (4) :} Summary: User1 argues that options A, C, and D are grammatically correct, but suggests that option C is the most appropriate due to its common usage in celebratory contexts. User2 counters that while options A and D are valid, option A is the most suitable because it expresses a positive willingness to help, making it the best choice overall.

Conclusion: Given the arguments, option A is more plausible as it conveys a positive sentiment and is grammatically correct.
\\

 \bottomrule
    \end{tabular}
    \caption{Qualitative example from M3-Exam.}
    \label{fig:qualitative_1}
\end{figure*}

\begin{figure*}[h]
    \centering
    \small
    \small
    \begin{tabular}{p{1.0\linewidth}}
    \toprule 
\textbf{(1) Native query:}  
Essa foi, provavelmente, a última Copa do Mundo de Futebol Feminino em que o público teve a chance de ver o trio de veteranas Marta, Cristiane e Formiga atuando juntas em campo. Formiga, que já chegou aos 41 anos e agora disputou sua sétima Copa, não deve voltar aos gramados para o mundial de 2023. No atual elenco da Seleção Feminina, por exemplo, são poucas as jogadoras que têm menos de 30 anos — sendo que, no masculino, essa faixa etária não é vista com tanta frequência entre os jogadores convocados para a Seleção.
Disponível em: https://capricho.abril.com.br. Acesso em: 3 set. 2019 (adaptado).
A longa permanência de atletas do futebol feminino na Seleção brasileira, numa faixa etária considerada avançada para o esporte, ocorre devido à \\
(A) pouca competitividade nos campeonatos.
(B) força do vínculo entre as atletas e a Seleção.
(C) capacidade de adaptação e resistência muscular.
(D) falta de incentivo para a modalidade.
\\
\midrule 
\textbf{(2) Translated query:}  
This was probably the last Women's World Cup in which the public had the chance to see the trio of veterans Marta, Cristiane and Formiga working together on the field. Formiga, who has reached the age of 41 and has now played in her seventh World Cup, is not expected to return to the pitch for the 2023 World Cup. In the current squad of the Women's National Team, for example, there are few players who are under 30 years old — and, in men, this age group is not seen as frequently among players called up for the national team.
Available at: https://capricho.abril.com.br. Accessed on: 3 Sep. 2019 (adapted).

The long stay of female football athletes in the Brazilian team, in an age group considered advanced for the sport, occurs due to the \\
(A) little competitiveness in championships.
(B) strength of the bond between the athletes and the National Team.
(C) adaptation capacity and muscular resistance.
(D) lack of incentive for the modality.
\\
\midrule 
\textbf{(3) Incorrect responses by \textsc{Eng-CoT}:} To analyze the question, we need to consider the context provided in the passage regarding the long careers of female athletes in the Brazilian national football team, particularly focusing on the older age of players like Formiga.

1. Option (A) - pouca competitividade nos campeonatos: This suggests that the lack of competitive opportunities might lead to players staying longer in the national team. However, this does not directly explain why older players are still selected; it rather implies a stagnation in player development.

2. Option (B) - força do vínculo entre as atletas e a Seleção: This option points to a strong emotional or professional connection between the players and the national team. This could be a valid reason for older players continuing to play, as they may feel a strong commitment to represent their country.

3. Option (C) - capacidade de adaptação e resistência muscular: This option suggests that older players have developed the ability to adapt and maintain their physical condition. While this could be true, it does not directly address the reason for their prolonged presence in the national team.

4. Option (D) - falta de incentivo para a modalidade: This implies that there is insufficient support for younger players to emerge, which could lead to older players remaining in the team longer. This could be a contributing factor, but it does not specifically highlight the emotional or professional ties that might keep veteran players in the squad.

Considering these points, option (B) seems to be the most relevant as it directly addresses the emotional and professional commitment of the players to the national team, which could explain why they continue to play at an older age.

Final answer: Answer: B
\\
\midrule 
\textbf{(4) Correct response by {Extracting}:} 
The passage highlights the presence of veteran players in the Brazilian women's national football team, particularly noting that many athletes are older than 30, which is considered an advanced age for competitive sports. This situation can be attributed to the lack of opportunities and resources in women's football, which often leads to fewer younger players being developed and promoted to the national team. As a result, the existing veteran players remain in the team longer due to the limited influx of younger talent. Therefore, the most appropriate answer is (D) falta de incentivo para a modalidade, as it reflects the systemic issues in women's football that contribute to the prolonged presence of older athletes in the national team.
\\
\midrule 
\textbf{(5) Example of \name{} response comparing (3) \& (4) :} Summary: The debate centers around the reasons for the long careers of older female athletes in the Brazilian national football team, particularly focusing on the presence of veteran players like Formiga. User1 argues that the strong emotional and professional bond between the players and the national team (Option B) is the primary reason for their continued participation. In contrast, user2 contends that the lack of opportunities and resources in women's football (Option D) leads to fewer younger players being developed, resulting in older players remaining in the team longer.

Conclusion: While both arguments present valid points, user2's perspective on the systemic issues in women's football (Option D) provides a broader context for understanding the situation. The lack of incentive and support for the development of younger talent is a significant factor that explains the prolonged presence of older athletes in the national team.
\\ \bottomrule
    \end{tabular}
    \caption{Qualitative example from M3-Exam.}
    \label{fig:qualitative_2}
\end{figure*}


\begin{table}[H]
\centering
\small
\caption{Human evaluation of \name{}}
\label{tab:human_eval_context}
\begin{tabular}{lc}
\toprule
\textbf{Metrics} & \textbf{Mean $\pm$ Std} \\
\midrule
Cultural Fidelity & $3.4 \pm 0.84$ \\
Factual Accuracy & $4.2 \pm 0.42$ \\
Stereotype/Bias & $1.9 \pm 1.10$ \\
Misinformation Risk & $1.7 \pm 0.82$ \\
\bottomrule
\end{tabular}
\end{table}

\section{Human Evaluation of Synthetic Context Quality}\label{app:human_eval}
To further assess the quality of \name{}-generated synthetic contexts, we conducted a focused human evaluation on the Korean subset of GlobalOpinionQA. 
We recruited five native Korean speakers who were also fluent in English, and asked them to evaluate the synthetic contexts generated by \name{} for 51 items. 
For each item, annotators were shown the question, the answer options, and the generated synthetic context, all in English, and rated the context on a 5-point Likert scale.

We considered four evaluation dimensions. First, \textit{Cultural Fidelity} measures whether the generated context aligns with Korean cultural and social perspectives. Second, \textit{Factual Accuracy} measures whether the background knowledge is correct and trustworthy. 
We also included two safety-related dimensions: \textit{Stereotype/Bias}, which captures whether the generated context contains culturally inappropriate generalizations or stereotypes, and \textit{Misinformation Risk}, which measures whether the context may mislead readers through inaccurate or unsupported claims.

Table~\ref{tab:human_eval_context} summarizes the results. 
\name{} achieves solid performance on both core quality dimensions, with a Cultural Fidelity score of $3.4 \pm 0.84$ and a Factual Accuracy score of $4.2 \pm 0.42$. In addition, the safety-related scores remain low, with Stereotype/Bias at $1.9 \pm 1.10$ and Misinformation Risk at $1.7 \pm 0.82$, indicating that the generated contexts rarely introduce problematic generalizations or misleading content.

Overall, these findings provide supporting evidence that \name{}'s synthetic contexts are not only culturally grounded and factually reliable, but also relatively safe in terms of bias and misinformation. 
Although this human study is limited to one language subset and a modest number of annotated examples, it offers an additional validation that the extracted context is generally appropriate for downstream multilingual reasoning.

\section{Usage of AI Assistants}
In preparing this work, we made limited use of AI-based writing assistants to improve sentence clarity, refine grammar, and enhance readability. 
These tools were employed strictly for editorial purposes and did not contribute to the conception of the research questions, development of methodology, execution of experiments, or interpretation of results. 
All technical content, analyses, and conclusions presented in this paper were conceived and authored solely by the researchers.

\end{document}